\documentclass{article}

\usepackage{microtype}
\usepackage{graphicx}
\usepackage{subcaption}
\usepackage{booktabs} 
\usepackage[most]{tcolorbox}
\usepackage{xcolor}
\usepackage{listings}

\usepackage{hyperref}

\usepackage[preprint]{icml2026}

\usepackage{amsmath}
\usepackage{amssymb}
\usepackage{mathtools}
\usepackage{amsthm}
\usepackage{enumitem}

\definecolor{mainblue}{RGB}{20, 30, 90}
\definecolor{maingreen}{RGB}{20, 80, 40}
\definecolor{backgray}{RGB}{245, 246, 250}

\definecolor{textgreen}{RGB}{46, 139, 87}
\definecolor{textred}{RGB}{200, 40, 40}
\definecolor{textorange}{RGB}{230, 120, 0}
\definecolor{textblue}{RGB}{0, 80, 200}
\definecolor{textgray}{RGB}{120, 120, 120}

\newtcolorbox{promptbox}[1][]{
  enhanced, 
  breakable,
  arc=4mm,
  colback=backgray,
  colframe=mainblue, 
  boxrule=1.5pt,
  fonttitle=\sffamily\bfseries,
  coltitle=white,        
  toptitle=3mm,           
  bottomtitle=3mm,        
  title=#1,                
  drop fuzzy shadow,       
}

\newtcolorbox{actionbox}[1][]{
  enhanced,
  breakable,
  arc=4mm,
  colback=backgray,
  colframe=maingreen,     
  boxrule=1.5pt,
  fonttitle=\sffamily\bfseries,
  coltitle=white,
  toptitle=3mm,
  bottomtitle=3mm,
  title=#1,
}

\usepackage[capitalize,noabbrev]{cleveref}

\theoremstyle{plain}

\theoremstyle{definition}

\theoremstyle{remark}

\usepackage[textsize=tiny]{todonotes}

\icmltitlerunning{G-LNS: Generative Large Neighborhood Search for LLM-Based Automatic Heuristic Design}

\begin{document}

\twocolumn[
  \icmltitle{G-LNS: Generative Large Neighborhood Search for \\LLM-Based Automatic Heuristic Design}



  \icmlsetsymbol{equal}{*}

  \begin{icmlauthorlist}
    \icmlauthor{Baoyun Zhao}{xxx}
    \icmlauthor{He Wang}{yyy,comp}
    \icmlauthor{Liang Zeng}{zzz}
  \end{icmlauthorlist}

  \icmlaffiliation{xxx}{Software College, Northeastern University, 110819, Shenyang, China}
  \icmlaffiliation{yyy}{International Centre for Theoretical Physics Asia-Pacific, University of Chinese Academy of Sciences, 100190, Beijing, China}
  \icmlaffiliation{comp}{Taiji Laboratory for Gravitational Wave Universe, University of Chinese Academy of Sciences, 100049, Beijing, China}
  \icmlaffiliation{zzz}{Tsinghua University, 100084, Beijing, China}

  \icmlcorrespondingauthor{He Wang}{hewang@ucas.ac.cn}
  \icmlcorrespondingauthor{Liang Zeng}{zengliangcs@gmail.com}

  \icmlkeywords{Large Language Models, Combinatorial Optimization, Large Neighborhood Search, Algorithm Discovery, Evolutionary Algorithms}

  \vskip 0.3in
]



\printAffiliationsAndNotice{}  

\begin{abstract}
While Large Language Models (LLMs) have recently shown promise in Automated Heuristic Design (AHD), existing approaches typically formulate AHD around constructive priority rules or parameterized local search guidance, thereby restricting the search space to fixed heuristic forms. Such designs offer limited capacity for structural exploration, making it difficult to escape deep local optima in complex Combinatorial Optimization Problems (COPs). In this work, we propose \emph{G-LNS}, a generative evolutionary framework that extends LLM-based AHD to the automated design of Large Neighborhood Search (LNS) operators. Unlike prior methods that evolve heuristics in isolation, G-LNS leverages LLMs to co-evolve tightly coupled pairs of \emph{destroy} and \emph{repair} operators. A cooperative evaluation mechanism explicitly captures their interaction, enabling the discovery of complementary operator logic that jointly performs effective structural disruption and reconstruction. Extensive experiments on challenging COP benchmarks, such as Traveling Salesman Problems (TSP) and Capacitated Vehicle Routing Problems (CVRP), demonstrate that G-LNS significantly outperforms LLM-based AHD methods as well as strong classical solvers. The discovered heuristics not only achieve near-optimal solutions with reduced computational budgets but also exhibit robust generalization across diverse and unseen instance distributions.\footnote{Our code are available at \url{https://github.com/zboyn/G-LNS}.}
\end{abstract}

\section{Introduction}

Combinatorial Optimization Problems (COPs) are ubiquitous in industrial manufacturing and logistics scheduling, where computational efficiency directly impacts operational costs \cite{dreo2006metaheuristics, desale2015heuristic}. Since many practical COPs are NP-hard \cite{garey2002computers}, hand-crafted heuristics have long been the dominant approach for solving large-scale instances \cite{blum2003metaheuristics, gendreau2010handbook}. However, traditional heuristic design relies heavily on domain expertise and is often tailored to specific problem structures, which substantially limits generalization across diverse tasks \cite{stutzle2018automated}.

\begin{figure}[tbp]
  \centering
  \includegraphics[width=\linewidth]{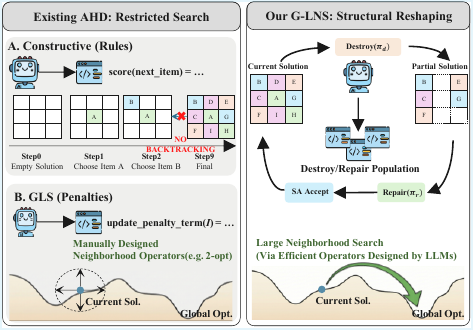}
  \caption{\textbf{Comparison of G-LNS and traditional AHD methods.} For combinatorial optimization problems, unlike existing AHD methods that are largely restricted to local search, G-LNS enables structural reshaping through LLM-generated LNS operators, allowing the search to escape local optima.}
  \label{fig:introduction}
  \vspace{-15pt}
\end{figure}

Recent advances in Large Language Models (LLMs), particularly in logical reasoning \cite{wei2022chain,zeng2024skywork,zhang2024sat} and code generation \cite{chen2021evaluating,zeng2025skywork}, have spurred growing interest in \emph{Automated Heuristic Design} (AHD) \cite{burke2013hyper}. By leveraging LLMs to automatically generate and refine algorithmic code, AHD searches for high-performance heuristics within a discrete algorithm space \cite{yang2023large}. Pioneering frameworks such as FunSearch \cite{romera2024mathematical} and EoH \cite{liu2024evolution} introduced the evolutionary ``Thought–Code" paradigm and demonstrated promising results on canonical tasks such as Bin Packing. Subsequent work extended this paradigm through reflective evolution \cite{ye2024reevo}, tree-based exploration strategies \cite{zheng2025monte}, and heuristic set evolution to improve generalization \cite{liu2025eoh}.

Despite this progress, existing AHD methods exhibit a fundamental \emph{structural bottleneck} (Figure~\ref{fig:introduction}). Most approaches instantiate AHD around either \emph{Constructive Heuristics} \cite{glover2001construction}, which evolve priority rules for sequential decision-making, or \emph{Guided Local Search} \cite{voudouris1996partial}, which optimizes penalty functions under fixed neighborhood operators. Constructive heuristics follow an irreversible trajectory: early suboptimal decisions are difficult to correct through later rule adjustments\cite{marti2011linear}. Conversely, while local search enables iterative refinement\cite{voudouris1999guided}, current AHD methods typically treat neighborhood structures (e.g., 2-opt\cite{johnson1997traveling}) as fixed priors, restricting the LLM to parameter tuning rather than enabling structural algorithmic innovation\cite{liu2024systematic}.

To overcome these limitations, we draw inspiration from \emph{Large Neighborhood Search} (LNS) \cite{shaw1998using}, a meta-heuristic that achieves strong structural reshaping through alternating \emph{destroy} and \emph{repair} operations\cite{ropke2006adaptive}. The effectiveness of LNS critically depends on the coupling between these two operators: the destroy phase determines the structural defects introduced into the solution, while the repair phase must be specifically adapted to reconstruct them\cite{pisinger2018large}. This strong interdependence makes automated LNS design particularly challenging and has largely prevented its adoption within existing AHD frameworks\cite{da2025large}.

In this work, we propose \emph{G-LNS}, an evolutionary framework that enables LLMs to automatically design problem-specific LNS operators. Instead of optimizing scalar heuristics or fixed templates, G-LNS prompts the LLM to generate executable code for both \emph{destroy} and \emph{repair} operators. To explicitly model their coupling, the framework maintains separate populations for destroy and repair operators and evaluates them jointly within an adaptive LNS process. A cooperative evaluation mechanism records the performance of operator pairs, allowing G-LNS to identify complementary destroy–repair logic and guide their co-evolution through synergy-aware crossover. This design allows the search process to move beyond local adjustments and discover heuristics capable of effective structural disruption and reconstruction. Our contributions are summarized as follows:

\begin{itemize}[leftmargin=*]
    \item \textbf{Generative LNS for AHD.} We propose G-LNS, an evolutionary framework that extends LLM-based AHD to the automated design of Large Neighborhood Search (LNS) operators, enabling structural solution perturbation beyond constructive rules and fixed local moves.

    \item \textbf{Synergy-aware Co-evolution.} We introduce a cooperative evaluation mechanism with a \emph{synergy matrix} to explicitly capture and exploit the coupling between \emph{destroy} and \emph{repair} operators during evolution.

    \item \textbf{Empirical Effectiveness and Generalization.} Extensive experiments on TSP and CVRP demonstrate that G-LNS outperforms state-of-the-art AHD methods and strong classical solvers, achieving near-optimal solutions with substantially reduced computational budgets.
\end{itemize}

\section{Background}
\subsection{Automatic Heuristic Design}
Automatic Heuristic Design (AHD) aims to automatically discover high-performance heuristics for combinatorial optimization problems (COPs)\citep{burke2013hyper, stutzle2018automated, chen2025heurigym}. Let $\mathcal{I}$ denote the instance space and $\mathcal{S}$ the solution space. A heuristic $h \in \mathcal{H}$ is a mapping $h: \mathcal{I} \rightarrow \mathcal{S}$ within a discrete algorithm space $\mathcal{H}$. The objective of AHD is to identify a heuristic $h^*$ that minimizes the expected objective value over a target distribution $\mathcal{D}$:
\begin{equation}
    h^* = \arg\min_{h \in \mathcal{H}} \ \mathbb{E}_{I \sim \mathcal{D}} \big[ f(h(I) \mid I) \big].
\end{equation}

While AHD enables systematic search over algorithmic logic, the expressiveness of the search space is determined by how heuristics are parameterized. Many existing approaches restrict the search to predefined templates, limiting the ability to induce fundamental changes in the underlying search dynamics. This work focuses on expanding the design space from fixed heuristic forms to structurally adaptive search operators.

\subsection{Large Neighborhood Search}
Large Neighborhood Search (LNS) is a meta-heuristic that explores the solution space through iterative \emph{destroy-and-repair} operations~\cite{shaw1998using}. Given a solution $x$, a destroy operator $d$ removes a subset of components (controlled by a destruction ratio $\epsilon$), producing a partial solution $x_{\text{partial}} = d(x)$. A repair operator $r$ then reconstructs a complete solution $x' = r(x_{\text{partial}})$.

To balance exploration and exploitation, LNS typically employs a Simulated Annealing (SA) acceptance criterion~\cite{henderson2003theory}, where non-improving solutions may be accepted with a probability that decreases over time. The performance of LNS critically depends on the coupling between $d$ and $r$: effective operators must introduce targeted structural disruption while enabling efficient reconstruction. Designing such complementary operator pairs remains a central challenge and motivates automation.

\subsection{Problem Formulation}
We formulate the automated design of LNS operators as an optimization problem over a discrete code space. Let $\Theta_d$ and $\Theta_r$ denote the spaces of executable code implementing destroy and repair operators, respectively. A policy $\pi = (d, r) \in \Theta_d \times \Theta_r$ is evaluated by running an LNS algorithm $\mathcal{A}$ with inherent stochasticity $\xi$ on instances $I \sim \mathcal{D}$.

The expected performance of $\pi$ is defined as
\begin{equation}
    J(\pi) = \mathbb{E}_{I \sim \mathcal{D}, \xi} \big[ f(\mathcal{A}(I, \pi; \xi) \mid I) \big],
\end{equation}
and the overall objective is to identify
\begin{equation}
    \pi^* = \arg\min_{\pi \in \Theta_d \times \Theta_r} J(\pi).
\end{equation}

\subsection{Related Work}
Recent work has integrated large language models (LLMs) into evolutionary frameworks for automated heuristic design~\cite{novikov2025alphaevolve, chen2025hifo}. Methods such as FunSearch~\cite{romera2024mathematical} and EoH~\cite{liu2024evolution} use LLMs as variation operators to generate heuristic code conditioned on prior implementations and performance feedback, forming a \emph{thought--code co-evolution} paradigm that improves exploration efficiency in large discrete algorithm spaces. However, most existing approaches rely on fixed algorithmic templates, focusing on constructive heuristics or parameter tuning within predefined local search frameworks\cite{ye2024reevo, zheng2025monte}, and thus largely overlook the structural design of neighborhood operators. In contrast, our work targets the automated generation of tightly coupled destroy-and-repair operators, enabling LLMs to reshape the search process at the structural level. Further discussion is provided in the appendix (Section~\ref{sec:related_works}).

\section{Methodology}
\label{sec:Methodology}
\begin{figure*}[htbp]
  \centering
  \includegraphics[width=\linewidth,trim=0 4mm 0 0mm,clip]{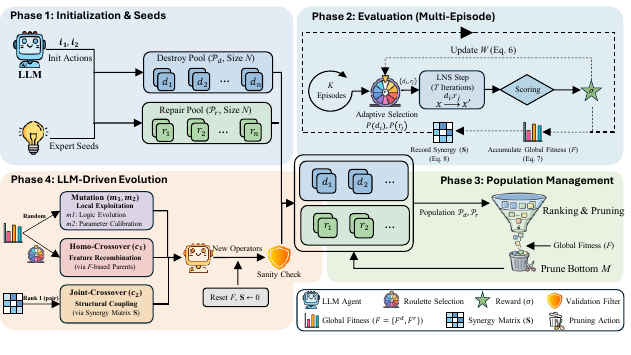}
  \caption{\textbf{The overall workflow of the G-LNS framework.} The framework operates in a cyclic manner consisting of four phases: (1) \textbf{Initialization}, where the dual populations ($\mathcal{P}_d, \mathcal{P}_r$) are seeded with domain expertise and LLM-generated operators; (2) \textbf{Evaluation}, where operator pairs are dynamically selected and scored via an Adaptive LNS process; (3) \textbf{Population Management}, which ranks and prunes low-performing operators; and (4) \textbf{Evolution}, leveraging LLMs to perform mutation and crossover strategies to replenish the population with novel heuristics.}
  \label{fig:G-LNS_overview}
  
\end{figure*}

\subsection{Overview of the G-LNS Framework}

We propose G-LNS, a framework for automating the discovery of high-performance \emph{destroy} and \emph{repair} operators in Large Neighborhood Search (LNS). G-LNS formulates heuristic design as an evolutionary process over algorithmic structures, where large language models (LLMs) act as intelligent variation operators to explore the space of algorithmic logic beyond fixed heuristic templates. This evolutionary formulation requires a reliable mechanism to assess the quality of newly generated operators.

To this end, we instantiate the LNS algorithm with an adaptive scoring mechanism inspired by Adaptive Large Neighborhood Search (ALNS)~\cite{ropke2006adaptive}, which provides a quantitative fitness signal by measuring each operator’s contribution to optimization performance. As illustrated in Figure~\ref{fig:G-LNS_overview}, the overall methodology consists of four phases—\textit{Initialization}, \textit{Evaluation}, \textit{Population Management}, and \textit{Evolution}.

\subsection{Initialization}

To facilitate the co-evolution of these interdependent components, G-LNS establishes a dual-population architecture\cite{potter1994cooperative}, maintaining distinct repositories for destroy operators ($\mathcal{P}_d$) and repair operators ($\mathcal{P}_r$), each with a capacity of $N$.
The initialization phase begins by injecting a compact set of classic domain-expert heuristics as seeds. For instance, in TSP/VRP tasks, we utilize Random Removal and Worst Removal for $\mathcal{P}_d$, and Greedy Insertion for $\mathcal{P}_r$.
These seeds serve a dual purpose: they provide foundational search capabilities and function as In-Context Examples\cite{brown2020language} to align the LLM with the specific task logic.
To fully populate the pools (when the number of seeds $< N$), we employ the Initialization Action (denoted as $i_1$ for destroy and $i_2$ for repair).
Through $i1, i2$, the LLM is prompted to conceptualize novel algorithmic logic and translate it into executable Python code, thereby ensuring the diversity of the search space from the onset.
(See Appendix~\ref{prompts_action} for specific details on prompt engineering).

Simultaneously, we initialize three metric structures to guide the evolutionary process. 
(1) \textbf{Global Fitness Scores} ($F=\{F^d, F^r\}$ with $F^d, F^r \in \mathbb{R}^N$): Initialized to zeros, these vectors track the cumulative performance of each operator in $\mathcal{P}_d$ and $\mathcal{P}_r$ for population management. 
(2) \textbf{Synergy Matrix} ($\mathbf{S} \in \mathbb{R}^{N \times N}$): Initialized to zeros, this matrix records the cooperative performance of specific pairs $(d_i, r_j)$ to guide joint crossover. 
(3) \textbf{Adaptive Weights} ($W=\{W^d, W^r\}$ with $W^d, W^r \in \mathbb{R}^N$): Initialized to ones, these weights are used solely within each evaluation episode to regulate the roulette wheel selection probabilities.

\subsection{Evaluation Phase}

The objective of this phase is twofold: to quantify the individual contribution of each operator for population management, and to capture the coupling effectiveness between destroy and repair operators for synergistic evolution. Given the inherent stochasticity of LNS, we employ a \textbf{Multi-Episode Evaluation Mechanism}.

We conduct $K$ independent evaluation episodes. Each episode starts from a random initial solution $x_0$ and executes for $T$ iterations. Importantly, the Adaptive Weights $W$ are reset to 1 at the beginning of each episode to ensure independent exploration, whereas the Global Fitness $F$ and Synergy Matrix $\mathbf{S}$ accumulate statistics across all $K$ episodes to obtain robust performance metrics that smooth out single-episode stochasticity.

\paragraph{Adaptive Selection.}

In each iteration $t$ of an episode, we select a destroy operator $d_i$ and a repair operator $r_j$ based on their current weights. The selection probability $P(d_i)$ (and similarly for $r_j$) follows the Roulette Wheel mechanism:
\begin{equation}
    P(d_i) = \frac{w^d_i}{\sum_{k=1}^{N} w^d_k}
\end{equation}

\paragraph{Scoring and Update.}

The selected pair $(d_i, r_j)$ is applied to the current solution to generate a neighbor $x'$. Its acceptance and the corresponding reward $\sigma$ are determined hierarchically.
If $x'$ improves the global best, we update both $x^*$ and $x_{\text{curr}}$ and assign $\sigma_1$.
If $x'$ only improves the current solution, we update $x_{\text{curr}}$ and assign $\sigma_2$.
For deteriorating solutions, acceptance is governed by the Metropolis criterion; if accepted, we update $x_{\text{curr}}$ with reward $\sigma_3$.
Otherwise, the solution is discarded and assigned the lowest reward ($\sigma_4$). Formally:
\begin{equation}
    \sigma = \begin{cases}
    \sigma_1, & \text{if } f(x') < f(x^*) \\
    \sigma_2, & \text{if } f(x') < f(x_{\text{curr}}) \\
    \sigma_3, & \text{if } x' \text{ is accepted by SA} \\
    \sigma_4, & \text{otherwise}
    \end{cases}
\end{equation}

Based on $\sigma$, we perform three distinct updates:

\begin{enumerate}
    \item \textbf{Adaptive Weights Update:} To guide the search direction during the current episode, the weights of the selected operators are updated dynamically using a smoothing factor $\lambda$:
    \begin{equation}
        w^d_i \leftarrow \lambda w^d_i + (1-\lambda) \sigma, \quad w^r_j \leftarrow \lambda w^r_j + (1-\lambda) \sigma
    \end{equation}
    
    \item \textbf{Global Fitness Accumulation:} To evaluate the overall quality of an operator for the subsequent \textit{Population Management} phase, we accumulate the reward into a global fitness score $F$:
    \begin{equation}
        F(d_i) \leftarrow F(d_i) + \sigma, \quad F(r_j) \leftarrow F(r_j) + \sigma
    \end{equation}
    
    \item \textbf{Synergy Recording:} To identify high-performing structural couplings, we update the corresponding entry in the synergy matrix:
    \begin{equation}
        \mathbf{S}_{ij} \leftarrow \mathbf{S}_{ij} + \sigma
    \end{equation}
    High values in $\mathbf{S}_{ij}$ indicate that destroy operator $i$ and repair operator $j$ possess complementary logic, which will be exploited in the \textit{Synergistic Joint Crossover} phase.
\end{enumerate}

\subsection{Population Management}
As the evolutionary process progresses, the operator pools $\mathcal{P}_d$ and $\mathcal{P}_r$ inevitably accumulate sub-optimal or redundant heuristics. Retaining these inefficient operators restricts the population's capacity to accommodate new, potentially superior logic.
Therefore, upon the completion of every $K$ evaluation episodes, we rank the operators in both populations based on their accumulated Global Fitness $F$. The system then prunes the bottom $M$ operators from each pool, thereby freeing up population slots for the subsequent \textit{LLM-driven Evolution} phase while preserving the high-performing elites.

\subsection{LLM-Driven Evolution Mechanism}
This phase leverages the code reasoning capabilities of LLMs to replenish the population slots vacated during management. We frame this as a heuristic search process within the algorithm space. To fill the $M$ empty slots, we randomly sample from three targeted evolutionary strategies, ensuring a diverse mix of local refinements and structural recombinations.

\textbf{Mutation (Local Exploitation) ($m_1, m_2$).} Mutation focuses on fine-tuning within the local algorithm space by modifying a single parent operator sampled from the elite pool. The specific action adapts to the parent's rank to balance stability and innovation: \textit{Logic Evolution ($m_1$)} is applied to lower-ranking elites to generate novel mechanisms for exploration; intermediate elites are assigned randomly; \textit{Parameter Calibration ($m_2$)} is applied to top-ranking elites to adjust hyperparameters (e.g., randomization ratios) for stability.

\textbf{Homogeneous Crossover (Feature Recombination) ($c_1$).} This strategy facilitates information exchange between operators of the same type. Two parents are selected via fitness-proportionate sampling based on $F$. The LLM is prompted to fuse the logical strengths of both parents, synthesizing a new operator that inherits hybrid characteristics (e.g., combining spatial clustering with random perturbation).

\textbf{Synergistic Joint Crossover (Structural Coupling) ($c_2$).} Addressing the inherent coupling between destroy and repair actions is a core innovation of G-LNS. We select the Destroy-Repair pair with the highest accumulated synergy score from $\mathbf{S}$. By conditioning the repair generation on the destroy logic, the LLM evolves this pair as a unified entity, ensuring the repair operator is specifically tailored to reconstruct the structural defects introduced by the destroy operator, thereby maximizing their synergistic performance.

\textbf{Robustness Guarantee.} To mitigate the risk of LLM hallucinations, we implement a \textit{Pre-evaluation Filter}\cite{chen2021evaluating}. All generated operators are subjected to a sanity check on a small-scale instance set\cite{romera2024mathematical}. Only those that are error-free and satisfy time complexity constraints are admitted to the population; otherwise, the regeneration process is triggered.

\textbf{State Reset.} Once the populations are fully replenished, the Global Fitness scores $F$ and the Synergy Matrix $\mathbf{S}$ are \textbf{reset to zero}. This ensures that the newly generated operators start on equal footing with the surviving elites in the subsequent evaluation cycle, preventing historical bias from dominating the search.

\section{Experiments}

To ensure a rigorously fair comparison, our experimental design strictly follows the experimental settings established in the aforementioned LLM-based AHD baselines. We evaluate G-LNS across three distinct domains: TSP, CVRP, and OVRP. Following these conventions, we utilize randomly generated instances for the evolution phase. During the testing phase, the learned operators are evaluated on both held-out generated instances and widely-used benchmark datasets (i.e., TSPLib\cite{reinelt1991tsplib} and CVRPLib\cite{uchoa2017new}) to verify their cross-distribution generalization. To mitigate the impact of randomness, we conduct three independent evolutionary runs for each task, capping the evolution process at $G_{max} = 200$ generations. This configuration represents a significant reduction compared to the 1,000 generations typically employed by the baselines, demonstrating that our framework achieves superior sample efficiency with a substantially lower token budget. In the final testing phase, the best-found operator pair is applied to test instances with a fixed budget of $T_{test}=500$ LNS iterations (increased from the $T=100$ iterations configured for the evolution phase) to rigorously measure solution quality. 

\textbf{Settings.} We employ \textit{DeepSeek-V3.2}\cite{liu2024deepseek} as the core LLM for operator generation. Regarding the evolutionary framework defined in Section \ref{sec:Methodology}, we maintain a population size of $N=5$ for both destroy and repair operator pools, and prune the bottom $M=2$ operators in each management phase.
For the evaluation process, we conduct $K=10$ independent episodes, each consisting of $T=100$ iterations. The hyperparameters for the inner LNS loop are configured as follows: initial temperature $T_0=100$, cooling rate $\alpha=0.97$, destruction ratio $\epsilon=0.2$, and weight update parameter $\lambda=0.5$. The scoring vector is set to $\sigma = \{1.5, 1.2, 0.8, 0.1\}$. Detailed descriptions of the evaluation datasets are provided in Appendix~\ref{details_dataset}. To ensure a fair comparison, all LLM-based AHD baselines utilize the same DeepSeek-V3.2 backbone.

\textbf{Baseline.} To verify the performance of G-LNS, we compare it against baselines from three categories: (1) \textbf{Handcrafted heuristics}. We employ LKH-3\cite{helsgaun2017extension} as the state-of-the-art baseline for TSP. For CVRP and OVRP, we utilize the widely-adopted solver OR-Tools to provide high-quality reference solutions. (2) \textbf{Neural Combinatorial Optimization (NCO)} methods, specifically POMO\cite{kwon2020pomo}, which serves as a representative constructive learning baseline for routing problems. (3) \textbf{LLM-based AHD methods}, including FunSearch\cite{romera2024mathematical}, EoH\cite{liu2024evolution}, ReEvo\cite{ye2024reevo}, EVO-MCTS\cite{wang2025automated}, and MCTS-AHD\cite{zheng2025monte}. Regarding the latter, we also evaluate its variant applied to Ant Colony Optimization (ACO), which evolves pheromone update rules and serves as a representative iterative AHD baseline. Additionally, we include a Standard ALNS equipped with classic domain-specific operators as a baseline to demonstrate the effectiveness of the generated operators\cite{ropke2006adaptive}. It is worth noting that while existing AHD methods primarily focus on evolving Constructive priority rules or tuning parameters for guided search, G-LNS extends the design space to the structural \textit{destroy-and-repair} logic of LNS.

\subsection{Experimental Results}
\begin{table*}[t]

\caption{Performance comparison on TSP and CVRP instances across five problem sizes. 
  \textbf{Top:} Results for TSP, where G-LNS is evolved on TSP50 and evaluated on 64 held-out instances for each size. Optimal solutions are derived using LKH-3. 
  \textbf{Bottom:} Results for CVRP ($C=50$) under the same evolution and evaluation protocol. Reference solutions are obtained using OR-Tools (total 320 s; 5 s per instance). 
  For all LLM-based AHD methods, reported values represent the average of three independent runs. 
  The overall best results are \underline{underlined}, and the best results among LLM-based AHD methods are highlighted in \textbf{bold}.}
  
  \label{tab:main-results}
  \begin{center}
  \begin{small}
  \setlength{\tabcolsep}{7.4pt}
      \begin{tabular}{l rrrrr rrrrr}
        \toprule
        \multicolumn{11}{c}{\textbf{Traveling Salesman Problem (TSP)}} \\
        \cmidrule(lr){1-11}
        & \multicolumn{2}{c}{TSP10} & \multicolumn{2}{c}{TSP20} & \multicolumn{2}{c}{TSP50} & \multicolumn{2}{c}{TSP100} & \multicolumn{2}{c}{TSP200} \\
        \cmidrule(lr){2-3} \cmidrule(lr){4-5} \cmidrule(lr){6-7} \cmidrule(lr){8-9} \cmidrule(lr){10-11}
        Method & Gap & Obj. & Gap & Obj. & Gap & Obj. & Gap & Obj. & Gap & Obj. \\
        \midrule
        Optimal (LKH-3) & \underline{0.00\%} & \underline{2.833} & \underline{0.00\%} & \underline{3.825} & \underline{0.00\%} & \underline{5.717} & \underline{0.00\%} & \underline{7.813} & \underline{0.00\%} & \underline{10.665} \\
        POMO            & \underline{0.00\%} & \underline{2.833} & 0.05\% & 3.827 & 0.43\% & 5.741 & 2.34\% & 7.996 & 20.35\% & 12.835 \\
        ALNS            & 0.25\% & 2.840 & 3.53\% & 3.960 & 5.65\% & 6.040 & 4.85\% & 8.192 & 5.96\% & 11.290 \\
        \midrule
        FunSearch       & 5.38\% & 2.986 & 11.78\% & 4.276 & 15.27\% & 6.590 & 17.24\% & 9.160 & 17.62\% & 12.544 \\
        EoH             & 4.73\% & 2.967 & 9.07\% & 4.172 & 14.73\% & 6.559 & 17.28\% & 9.163 & 17.84\% & 12.568 \\
        ReEvo           & 3.15\% & 2.922 & 6.97\% & 4.092 & 10.86\% & 6.338 & 12.88\% & 8.820 & 14.77\% & 12.240 \\
        MCTS-AHD        & 2.85\% & 2.914 & 7.84\% & 4.125 & 11.24\% & 6.359 & 12.02\% & 8.753 & 13.16\% & 12.068 \\
        Evo-MCTS        & 2.09\% & 2.892 & 4.96\% & 4.015 & 7.82\% & 6.164 & 9.54\% & 8.599 & 10.20\% & 11.753 \\
        MCTS-AHD(ACO)   & 0.11\% & 2.836 & 0.27\% & 3.836 & 1.21\% & 5.786 & 3.45\% & 8.083 & 6.22\% & 11.329 \\
        \textbf{Ours}   & \textbf{0.00\%} & \textbf{2.833} & \textbf{0.01\%} & \textbf{3.826} & \textbf{0.37\%} & \textbf{5.738} & \textbf{1.10\%} & \textbf{7.899} & \textbf{1.31\%} & \textbf{10.805} \\
        
        \midrule[\heavyrulewidth]
        
        \multicolumn{11}{c}{\textbf{Capacitated Vehicle Routing Problem (CVRP, $C=50$)}} \\
        \cmidrule(lr){1-11}
        & \multicolumn{2}{c}{CVRP10} & \multicolumn{2}{c}{CVRP20} & \multicolumn{2}{c}{CVRP50} & \multicolumn{2}{c}{CVRP100} & \multicolumn{2}{c}{CVRP200} \\
        \cmidrule(lr){2-3} \cmidrule(lr){4-5} \cmidrule(lr){6-7} \cmidrule(lr){8-9} \cmidrule(lr){10-11} 
        Method & Gap & Obj. & Gap & Obj. & Gap & Obj. & Gap & Obj. & Gap & Obj. \\
        \midrule
        OR-Tools        & \underline{0.00\%} & \underline{3.096} & \underline{0.00\%} & \underline{4.606} & \underline{0.00\%} & \underline{8.145} & 2.09\% & 14.106 & 1.27\% & 25.088 \\
        POMO            & 2.42\% & 3.171 & 3.81\% & 4.781 & 4.27\% & 8.493 & 4.69\% & 14.466 & 28.11\% & 31.738 \\
        ALNS            & 2.45\% & 3.172 & 4.62\% & 4.818 & 4.26\% & 8.492 & 3.43\% & 14.291 & 3.63\% & 25.674 \\
        \midrule
        FunSearch       & 11.94\% & 3.466 & 26.34\% & 5.819 & 33.49\% & 11.014 & 31.75\% & 18.205 & 24.87\% & 30.936 \\
        EoH             & 11.31\% & 3.446 & 21.38\% & 5.591 & 26.88\% & 10.468 & 27.27\% & 17.585 & 19.50\% & 29.606 \\
        ReEvo           & 11.10\% & 3.440 & 22.00\% & 5.619 & 24.94\% & 10.308 & 24.42\% & 17.192 & 17.60\% & 29.135 \\
        MCTS-AHD        & 11.10\% & 3.440 & 20.65\% & 5.557 & 23.88\% & 10.221 & 23.77\% & 17.102 & 16.37\% & 28.831 \\
        Evo-MCTS        & 8.07\% & 3.346 & 17.52\% & 5.412 & 21.67\% & 10.038 & 19.50\% & 16.512 & 12.79\% & 27.942 \\
        MCTS-AHD(ACO)   & 2.80\% & 3.183 & 6.85\% & 4.921 & 11.65\% & 9.094 & 12.90\% & 15.600 & 10.52\% & 27.380 \\
        \textbf{Ours}   & \textbf{1.44\%} & \textbf{3.141} & \textbf{2.20\%} & \textbf{4.707} & \textbf{1.29\%} & \textbf{8.250} & \underline{\textbf{0.00\%}} & \underline{\textbf{13.817}} & \underline{\textbf{0.00\%}} & \underline{\textbf{24.774}} \\
        \bottomrule
      \end{tabular}
  \end{small}
  \end{center}
  \vskip -0.1in
\end{table*}

\textbf{Performance on synthetic held-out instances.}
As evidenced in Table \ref{tab:main-results} (Top), \textbf{G-LNS (Ours)} consistently achieves the lowest optimality gaps among all LLM-driven methods across all problem sizes on TSP tasks. For large-scale instances such as TSP100 and TSP200, our framework significantly outperforms both Evo-MCTS and the iterative MCTS-AHD(ACO), effectively overcoming the scalability challenges that often lead existing AHD methods to exhibit gaps exceeding 10\%. 
This superior performance is driven by the evolution of state-dependent destroy and repair logic, which allows G-LNS to outperform Standard ALNS detailed in Appendix~\ref{app:designed_op}. Specifically, it identifies \textit{Adaptive Continuous-Segment Removal} and \textit{Diversity-Aware Insertion} for TSP, and \textit{Progressive Stochastic-Worst Removal} paired with \textit{Context-Aware Greedy Insertion} for CVRP.
Unlike static rules, these operators dynamically adjust destruction magnitude and exploration noise based on real-time solution states, enabling superior escape from local optima.

On CVRP tasks (Table \ref{tab:main-results}, Bottom), G-LNS exhibits remarkable scalability and robustness in the face of complex capacity constraints. These constraints often pose significant challenges for traditional constructive heuristics, which tend to get trapped in local optima due to the irreversible nature of their sequential decisions. While G-LNS performs slightly below the optimal reference solutions on small-scale instances, its advantages become more pronounced as the problem complexity increases. On the largest instances (CVRP100/200), G-LNS not only outperforms all LLM-based baselines—which typically show optimality gaps exceeding 10\%—but also identifies solutions superior to those provided by the OR-Tools solver. The ability to navigate this constrained solution space suggests that the learned destroy-and-repair operators can successfully correct structural defects that constructive methods are unable to address.(See Appendix~\ref{app:ovrp_details} for details on OVRP).

G-LNS achieves superior solution quality while exhibiting significantly higher computational efficiency compared to benchmark methods. In CVRP experiments, while the OR-Tools baseline utilizes a fixed budget of 320 seconds for the evaluation batch (5 seconds per instance for 64 instances), G-LNS requires substantially less time across all problem scales: its total inference time for 500 iterations ranges from merely 3.23s for CVRP10 to 280.91s for CVRP200. In contrast, MCTS-AHD, another iterative approach, incurs a much heavier computational burden, requiring 84.16s for CVRP10 and 2407.14s for CVRP200, making G-LNS over an order of magnitude faster. These results demonstrate that G-LNS can identify higher-quality solutions than both the standard solver—which fails to fully converge within the time limit and leaves a gap of 1.27\%--2.09\% —and expensive iterative heuristics, all while consuming a much smaller computational budget.

\textbf{Robust generalization to real-world benchmarks.}
To further assess the cross-distribution generalization capability of G-LNS, we extended our evaluation to widely recognized standard benchmarks, including TSPLib\cite{reinelt1991tsplib} and CVRPLib\cite{uchoa2017new}. These datasets feature diverse problem distributions and scales that differ significantly from the random instances used during evolution.

As detailed in Appendix~\ref{app:benchmark_details}, G-LNS demonstrates superior generalization performance compared to state-of-the-art AHD methods, including the strong baseline EoH-S~\citep{liu2025eoh}. G-LNS consistently achieves the lowest optimality gaps across all evaluated datasets. Notably, on the challenging CVRPLib Set F, our method reduces the optimality gap from 40.1\% (EoH-S) to \textbf{15.9\%}. Similarly, on TSPLib, G-LNS maintains a low gap of \textbf{2.8\%}, significantly outperforming baselines that struggle with unseen distributions. These results confirm that the destroy-and-repair operators evolved by G-LNS capture intrinsic structural properties of combinatorial problems rather than merely overfitting to the training distribution.

\subsection{Ablation Studies}

To validate the necessity of each evolutionary strategy and component within the G-LNS framework, we conducted ablation studies on the TSP50 and CVRP50 dataset. We established the full G-LNS as the baseline and compared it against four degenerated variants:

\textbf{w/o Mut. (No Mutation)} excludes the local refinement strategies (i.e., Logic Evolution and Parameter Calibration). The evolution relies entirely on crossover operations, assessing the impact of fine-tuning single operators.
\textbf{w/o Homo. (No Homogeneous Crossover)} removes the mechanism of fusing operators of the same type. New operators are generated solely through mutation or synergistic pairing, testing the benefit of recombining high-level logic features within the same operator class.
\textbf{w/o Syn. (No Synergistic Joint Crossover)} decouples the evolution of destroy and repair operators. Instead of evolving complementary pairs based on the synergy matrix $\mathbf{S}$, it treats the populations independently.
\textbf{w/o Adapt. (No Adaptive Weights)} disables the Adaptive LNS scoring mechanism during the evaluation phase, where operators are selected with uniform probability to verify the importance of dynamic resource allocation.

\begin{table}[h]
  \caption{Ablation study of the key components in G-LNS. Each variant removes one component—Mutation (Mut.), Homogeneous Crossover (Homo.), Synergistic Crossover (Syn.), or Adaptive Weights (Adapt.)—to assess its impact on solution quality for TSP50 and CVRP50. The best results are highlighted in \textbf{bold}.}
  \label{tab:ablation-results}
  \begin{center}
    \begin{small}
        \setlength{\tabcolsep}{15pt}
        \begin{tabular}{lcc}
          \toprule
          & TSP50 & CVRP50 \\
          \midrule
          \textbf{G-LNS (Original)} & \textbf{0.37\%} & \textbf{1.29\%} \\
          ALNS                  & 5.65\%          &   4.26\% \\
          \midrule
          \textit{w/o} Mut.              & 1.55\%          &  1.96\% \\
          \textit{w/o} Homo.             & 1.40\%          &  2.03\% \\
          \textit{w/o} Syn.              & 1.24\%          &  1.87\% \\
          \textit{w/o} Adapt.            & 0.95\%          &  1.68\% \\
          \midrule
          G-LNS Flat                     & 1.69\%          & 2.31\% \\
          G-LNS Aggressive               & 0.51\%          & 1.50\% \\
          \bottomrule
        \end{tabular}
    \end{small}
  \end{center}
\end{table}

Furthermore, to evaluate the robustness of the adaptive mechanism, we conducted a sensitivity analysis on the scoring vector $\sigma = \{\sigma_1, \sigma_2, \sigma_3, \sigma_4\}$ which governs operator weight updates. We compared our default setting against a \textbf{Flat} configuration ($\{1, 1, 1, 0.1\}$), where rewards for different success levels are indistinguishable, and an \textbf{Aggressive} configuration ($\{10, 5, 2, 0\}$), amplifying the reward variance to verify the necessity of a hierarchical reward system.

Table \ref{tab:ablation-results} shows G-LNS achieves the lowest gaps, validating our evolutionary strategies.
(1) \textbf{Evolutionary Components:} Significant drops in \textbf{w/o Mut.} and \textbf{w/o Homo.} highlight the importance of logic fine-tuning and feature recombination. The degradation in \textbf{w/o Syn.} confirms the structural coupling between destroy and repair operators; independent evolution disrupts their synergy.
(2) \textbf{Scoring Mechanism:} The decline in \textbf{w/o Adapt.} verifies the feedback loop's value. Notably, the \textbf{Flat} vector performs worse than removing adaptivity entirely, proving that indistinguishable rewards mislead the search. Meanwhile, the \textbf{Aggressive} setting falls short of the default, confirming the necessity of a balanced hierarchical reward system.

\subsection{Convergence Analysis}

\begin{figure}[t]
  \centering
  \includegraphics[width=\linewidth]{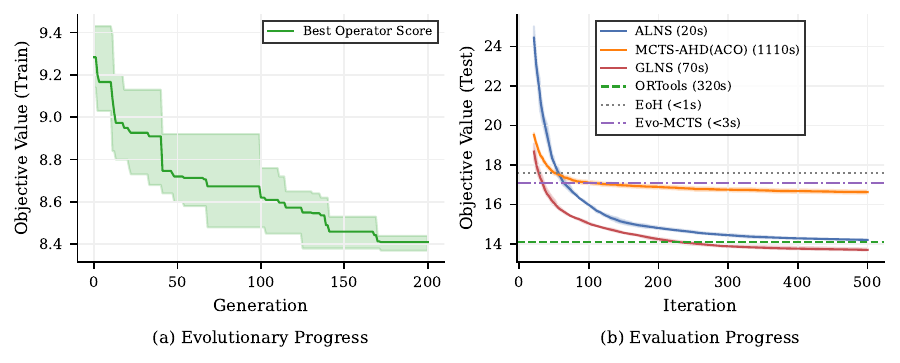}
  \caption{\textbf{Convergence and Evolutionary Analysis.} \textbf{(a) Evolutionary Progress:} Validation score trajectory of the best operator over 200 generations; the steady decline confirms the LLM's capacity to evolve high-performance heuristics. \textbf{(b) Evaluation Progress:} Convergence comparison on CVRP100 instances. G-LNS identifies the best solution in 70s across all 64 instances, significantly outperforming both the Solver (320s) and MCTS-AHD(ACO) (1110s) in terms of search efficiency.}
  \label{fig:convergence}
  \vspace{-20pt}
\end{figure}

Figure \ref{fig:convergence} illustrates the dual performance characteristics of G-LNS: the progressive improvement of operator quality during evolution (Fig. \ref{fig:convergence}a) and the rapid convergence of the final evolved operators during evaluation (Fig. \ref{fig:convergence}b).

\textbf{Evolutionary Efficiency.} Figure \ref{fig:convergence}(a) depicts the validation performance of the elite operators across 200 generations. We observe a rapid quality improvement in the initial phase (Generations 0-50), indicating that the LLM quickly grasps the fundamental logic of destroy-and-repair operations from the seed examples. Subsequently, the curve exhibits a steady refinement trend (Generations 50-200), where the framework fine-tunes the operator logic to escape local optima. The narrowing variance (shaded area) suggests that the population converges towards a set of robust and high-performing heuristics.

\textbf{Solving Convergence.} Figure \ref{fig:convergence}(b) compares the convergence behavior of G-LNS against representative baselines on CVRP100 instances. Several key observations can be drawn:
(1) \textbf{Superior Convergence Rate:} Compared to Standard ALNS and MCTS-AHD(ACO), G-LNS demonstrates a significantly steeper descent in the early iterations. This suggests that the LLM-generated destroy operators possess stronger structural perturbation capabilities, allowing the search to quickly identify promising regions in the solution space.
(2) \textbf{Beating the Solver:} G-LNS surpasses the solution quality of constructive baselines within the first 50 iterations. More importantly, it eventually converges to a solution ($Obj \approx 13.8$) superior to that of the OR-Tools solver ($Obj \approx 14.1$).
(3) \textbf{Computational Efficiency:} G-LNS achieves state-of-the-art performance in approximately 70 seconds, which is not only 4.5$\times$ faster than the 320-second budget allocated to OR-Tools , but also orders of magnitude more efficient than MCTS-AHD(ACO), which requires 1110 seconds. This dramatic reduction in computational overhead highlights the practical value of our evolved heuristics for time-critical applications.

\subsection{Case Study}

\begin{figure}[t]
    \centering
    \captionsetup[subfigure]{font=scriptsize, labelfont=scriptsize}
    \begin{subfigure}[b]{0.32\linewidth}
        \includegraphics[width=\linewidth, trim=0 0 0 23pt, clip]{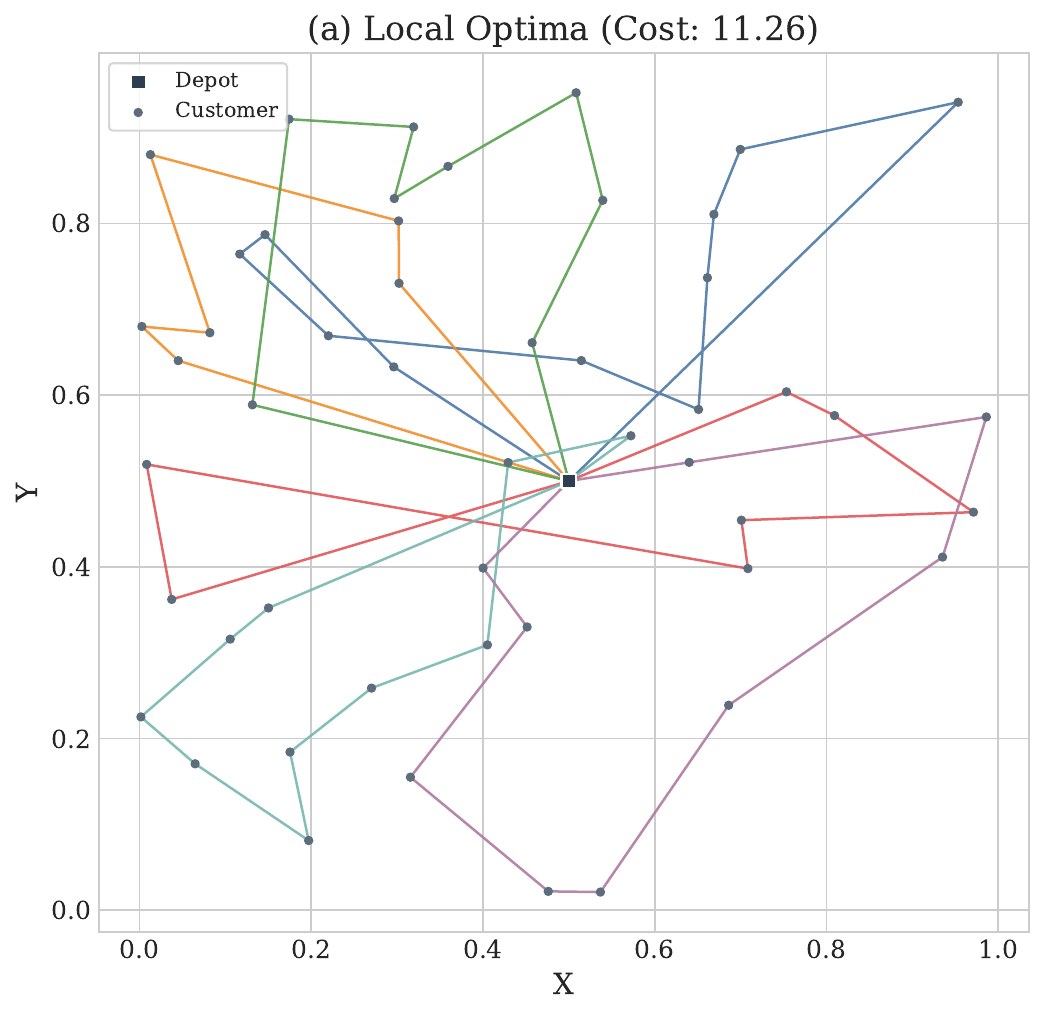}
        \caption{Before ($C=11.26$)}
        \label{fig:case_before}
    \end{subfigure}
    \hfill
    \begin{subfigure}[b]{0.32\linewidth}
        \includegraphics[width=\linewidth, trim=0 0 0 23pt, clip]{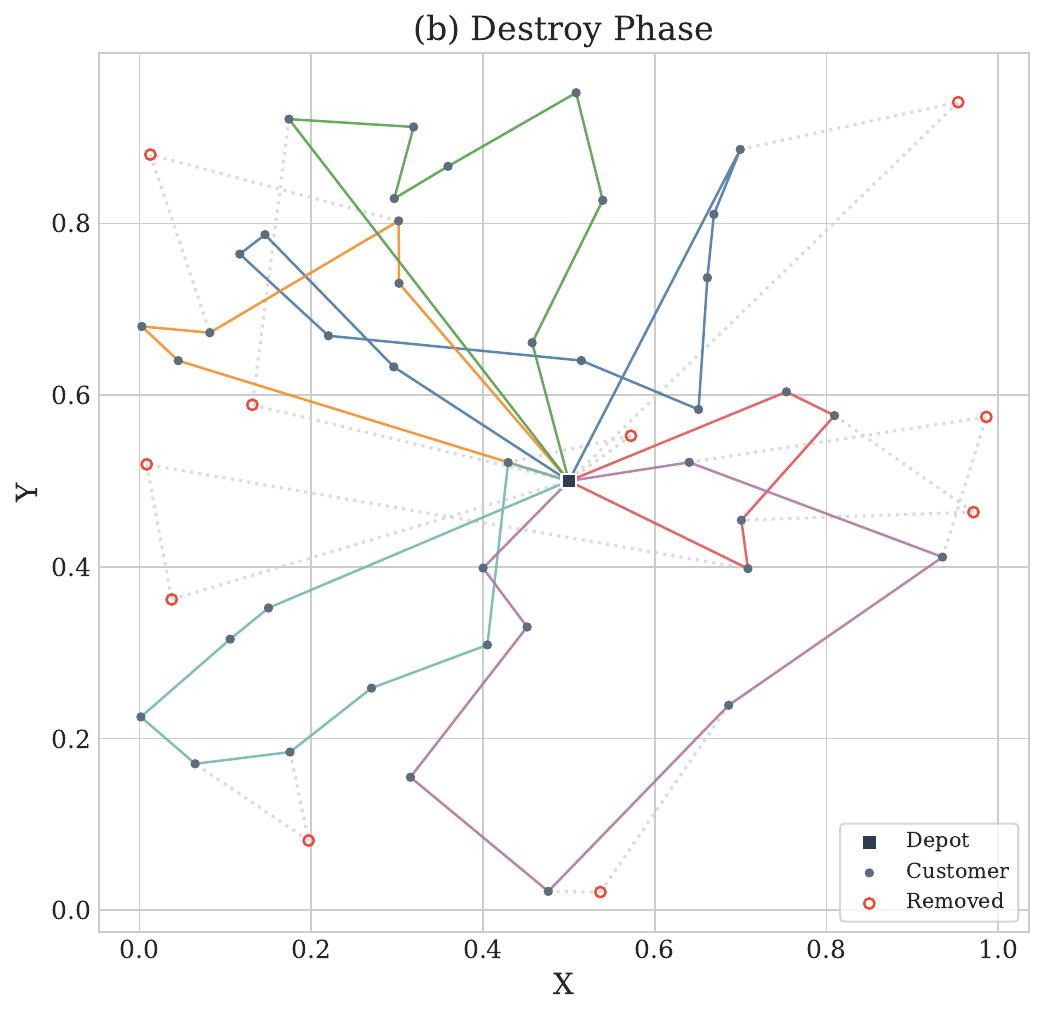}
        \caption{Destroy}
        \label{fig:case_destroy}
    \end{subfigure}
    \hfill
    \begin{subfigure}[b]{0.32\linewidth}
        \includegraphics[width=\linewidth, trim=0 0 0 23pt, clip]{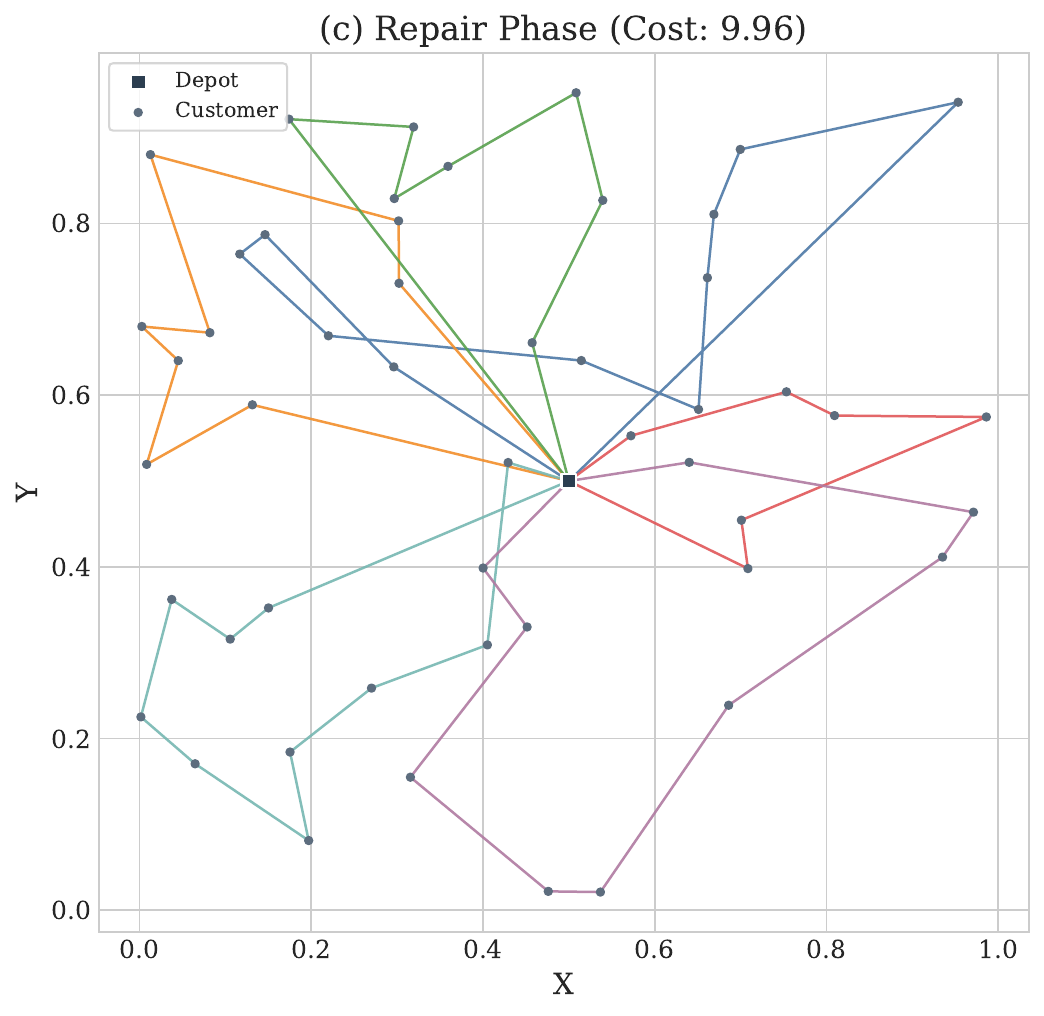}
        \caption{Repair ($C=9.96$)}
        \label{fig:case_repair}
    \end{subfigure}
    \caption{\textbf{Case Study on Structural Reshaping.} Visualizing a snapshot of the evolutionary process on CVRP50. \textbf{(a-b)} The generated repair operator targets the entangled region for destruction. \textbf{(c)} The destroy operator resolves the crossing by optimizing node-to-vehicle assignments, reducing the cost from 11.26 to 9.96.}
    \label{fig:case_study_narrow}
    \vspace{-20pt}
\end{figure}

Figure~\ref{fig:case_study_narrow} illustrates a representative snapshot of a single optimization iteration during the evolutionary process of G-LNS on a CVRP50 instance.
The iteration begins with a solution trapped in a local optimum ($Cost=11.26$, Fig.~\ref{fig:case_before}), characterized by crossing edges and inefficient routings.

In the Destroy phase (Fig.~\ref{fig:case_destroy}), applying a destruction rate of $\epsilon=0.2$, the evolved \textit{PSWR} operator (Appendix~\ref{app:op_cvrp}) exhibits a targeted strategy. 
It specifically identifies and removes nodes involved in the most entangled regions, effectively dismantling the sub-optimal structures to facilitate re-optimization. Subsequently, the \textit{ACAGI} repair operator (Fig.~\ref{fig:case_repair}) reconstructs the solution. Crucially, this process goes beyond merely re-sequencing nodes within their original paths. As shown in the transition from (b) to (c), the operator dynamically reassigns customers to different routes, correcting sub-optimal node-to-vehicle assignments. This simultaneous optimization of clustering and sequencing successfully untangles the crossings, significantly reducing the cost to $9.96$.

\section{Conclusion}
In this paper, we introduced G-LNS, a generative evolutionary framework that overcomes the structural limitations of existing LLM-based Automated Heuristic Design by co-evolving tightly coupled destroy and repair operators. Through a synergy-aware evaluation mechanism and novel crossover strategies, G-LNS successfully discovers sophisticated heuristic logic that significantly outperforms state-of-the-art baselines and strong classical solvers on complex routing problems, while demonstrating superior generalization capabilities. For future work, we plan to extend G-LNS to multi-objective optimization settings and investigate its applicability to a broader range of combinatorial problems beyond routing tasks.

\section{Acknowledge}
We express our gratitude to the team behind the LLM4AD platform\cite{liu2024llm4ad}. Their open-source framework significantly facilitated the implementation of baseline methods and provided a robust environment for our comparative experiments.
We also thank the DeepSeek team for developing the DeepSeek-V3.2 model\cite{liu2024deepseek}, which served as the core LLM in our framework.
This work was supported by the National Key Research and Development Program of China under Grant No. 2021YFC2203004. HW acknowledges support from the National Natural Science Foundation of China (NSFC) under Grant Nos. 12547104 and 12405076.




\bibliography{example_paper}
\bibliographystyle{icml2026}

\newpage
\appendix
\onecolumn

\section{More Discussion on Related Work}
\label{sec:related_works}
\subsection{AHD}

Automated Heuristic Design (AHD), frequently referred to as Hyper-heuristics \citep{burke2013hyper}, aims to automate the discovery of optimization algorithms. Formally, instead of searching for an optimal solution in the solution space $\mathcal{S}$, AHD searches for a high-quality heuristic algorithm $h$ within an algorithm space $\mathcal{H}$. The objective is to identify heuristics that generalize well across a target distribution of problem instances, rather than overfitting to a single case.

Historically, Evolutionary Computation (EC) has served as the primary search strategy for AHD. Among various EC methods\cite{lopez2016irace,blot2016mo,burke2018classification}, Genetic Programming (GP)\cite{koza1994genetic, o2009riccardo} has long been considered the prevailing approach. In the GP paradigm, heuristics are typically represented as syntax trees, which are evolved through genetic operations such as subtree crossover and point mutation to optimize their performance on training instances\citep{mei2022explainable}.

Despite its success, traditional GP-based AHD faces a significant bottleneck: the reliance on hand-crafted genetic operators. The mutation and crossover operators often require substantial domain expertise to ensure that the modified heuristics remain syntactically valid and semantically meaningful \citep{duflo2019gp}. This dependency on manual design limits the flexibility of the search process, motivating the recent shift towards more intelligent, generative approaches for algorithm discovery.

\subsection{Neural Combinatorial Optimization (NCO)}

Neural Combinatorial Optimization (NCO) has emerged as a promising paradigm to address the computational prohibitiveness of traditional exact solvers. The core motivation of NCO is to learn heuristics from data offline, enabling the generation of high-quality approximate solutions in real-time inference \citep{bello2016neural, bengio2021machine}.

\textbf{Sequence-to-Sequence Modeling.}
Early NCO approaches formulated the construction of solutions as a sequence-to-sequence (Seq2Seq) prediction task, similar to neural machine translation \citep{sutskever2014sequence}. However, standard Seq2Seq models struggled with combinatorial problems because the output vocabulary (e.g., the specific cities to visit) varies for each input instance, unlike the fixed vocabulary in translation tasks \citep{vinyals2015pointer}. To overcome this limitation, \citet{vinyals2015pointer} introduced the \textit{Pointer Network} (Ptr-Net), which employs a pointer mechanism to select input elements directly as outputs using attention scores.

\textbf{From Supervised Learning to RL.}
While Ptr-Nets laid the foundation, optimizing them via Supervised Learning proved impractical due to the high cost of obtaining optimal labels for large-scale instances \citep{bello2016neural}. Consequently, the field shifted towards Reinforcement Learning (RL), treating the generation process as a Markov Decision Process \citep{nazari2018reinforcement}. \citet{bello2016neural} proposed an Actor-Critic framework where the neural network acts as a policy to minimize the tour length, using the negative tour length as the reward signal. They also introduced Active Search to refine solutions during inference by sampling multiple trajectories to find the best candidate \citep{bello2016neural}.

\textbf{Symmetry-Aware Transformer (POMO).}
Modern NCO methods have evolved to leverage the Transformer architecture for better feature extraction and long-range dependency modeling \citep{kool2018attention,bresson2021transformer}. Notably, POMO \citep{kwon2020pomo} addressed a critical start node bias inherent in previous constructive policies. By exploiting the rotational symmetry of routing problems, POMO generates diverse trajectories in parallel (one for each starting node) and uses the average reward as a low-variance \textit{shared baseline} \citep{kwon2020pomo}. This approach significantly stabilizes training and achieves state-of-the-art performance among constructive NCO methods.

\textbf{Limitations and The Shift to AHD.}
Despite their fast inference speed, NCO models fundamentally operate as black boxes and often suffer from poor generalization \citep{bengio2021machine}. Their performance typically degrades significantly when applied to problem scales or distributions unseen during training (distributional shift) \citep{joshi2020learning, fu2021generalize}. These limitations regarding interpretability and scalability have motivated the recent surge in \textit{LLM-based Automated Heuristic Design}, which aims to evolve explicit, generalizable algorithmic code instead of opaque neural weights \citep{romera2024mathematical, liu2024evolution}.

\subsection{LLM for Combinatorial Optimization}
Recent advancements have spurred a paradigm shift in applying Large Language Models (LLMs) to Combinatorial Optimization (CO). We categorize existing methodologies into two distinct streams: \textit{LLM as Solver} and \textit{LLM as Designer}.

\textbf{LLM as Solver (Direct \& Iterative Optimization).}
This paradigm treats LLMs as black-box optimizers, prompting them to output solutions directly based on problem descriptions. Early attempts employed zero-shot or few-shot prompting to solve small-scale instances \citep{guo2023connecting}. To improve performance, \citet{yang2023large} introduced \textit{Optimization by PROmpting (OPRO)}, where the LLM iteratively refines solutions using natural language feedback and optimization trajectories as in-context information. Other works explore fine-tuning LLMs specifically for CO tasks to enhance their understanding of constraints \citep{jiang2024unco}.

However, the LLM as Solver paradigm faces intrinsic limitations. First, LLMs struggle with the \textit{tokenization of high-precision coordinates} and numerical reasoning, often perceiving numbers as linguistic tokens rather than mathematical values \citep{wu2024pre}. Second, as noted by \citet{kambhampati2024llms}, LLMs function better as idea generators than reliable planners; they lack the rigorous backtracking capabilities required for NP-hard problems. Consequently, these methods are prone to hallucinations and scale poorly to large instances due to limited context windows.

\textbf{LLM as Designer (Automated Heuristic Design).}
Acknowledging the limitations of direct inference, the field has gravitated towards \textit{LLM-based AHD}, repurposing LLMs to generate executable code. This paradigm ensures correctness and scalability by offloading execution to standard Python interpreters. Pioneering frameworks like FunSearch \citep{romera2024mathematical} and EoH \citep{liu2024evolution} established the foundational Thought-Code evolutionary paradigm, treating LLMs as mutation operators to evolve populations of constructive heuristics\cite{huang2025calm}.

\textit{Advanced Search Strategies.}
To overcome the tendency of standard population-based methods to converge into local optima, recent works have introduced more sophisticated search mechanisms. ReEvo \citep{ye2024reevo} integrates a reflective evolution mechanism, mimicking human thinking to retrospectively analyze historical performance and guide more effective code mutations. Furthermore, MCTS-AHD \citep{zheng2025monte} and Evo-MCTS \citep{wang2025automated} introduce Monte Carlo Tree Search (MCTS) into the evolutionary process. By organizing heuristics in a tree structure, these methods balance exploration and exploitation, allowing for the comprehensive development of temporarily underperforming heuristics that standard populations might prematurely discard.

\textit{Heuristic Set Evolution.}
Addressing the generalization bottleneck where a single heuristic fails to cover diverse instance distributions, EoH-S \citep{liu2025eoh} proposes the concept of \textit{Automated Heuristic Set Design (AHSD)}. Instead of optimizing a solitary algorithm, EoH-S evolves a complementary set of heuristics, ensuring that each problem instance is covered by at least one effective strategy in the set, thereby achieving superior cross-distribution performance.

\textit{Differentiation from LLM-driven Heuristic Neighborhood Search.}
It is crucial to distinguish our approach from the recently proposed LHNS \citep{xie2025llm}. LHNS applies the logic of neighborhood search to the \textit{algorithm space} itself—iteratively perturbing heuristic code blocks to find better functions. While LHNS uses LNS-like concepts to guide the code search process, the output remains a standard heuristic function. 

\textbf{Summary: A Paradigm Shift in Framework.}
As summarized in Table \ref{tab:related_works_comparison}, in contrast to prior arts, our G-LNS does not merely evolve a better scoring function or a local guidance rule; it fundamentally alters the algorithmic framework. By explicitly tasking the LLM with designing structural \emph{destroy} and \emph{repair} operators, G-LNS enables the solver to perform complex topological transformations on the solutions. This represents a shift from optimizing parameters/rules within a fixed skeleton to automating the design of the solver's structural components, a capability that extends beyond the scope of previous AHD methods.

\begin{table*}[t]
\centering
\caption{Comparison of LLM-based approaches for Combinatorial Optimization. Our G-LNS is unique in targeting the structural design of LNS operators.}
\label{tab:related_works_comparison}
\resizebox{\textwidth}{!}{%
\begin{tabular}{l|c|c|c|l}
\toprule
\textbf{Method} & \textbf{LLM Role} & \textbf{Target Heuristic Type} & \textbf{Search Strategy} & \textbf{Key Characteristic / Focus} \\ 
\midrule
\multicolumn{5}{c}{\textit{LLM as Solver (Direct Inference)}} \\
\midrule
\textbf{OPRO} \citep{yang2023large} & Solver & N/A (Direct Solution) & Iterative Prompting & Optimizes solutions via natural language feedback history. \\
\textbf{Fine-tuned LLMs} \citep{jiang2024unco} & Solver & N/A (Direct Solution) & Supervised Fine-tuning & Enhances LLM's domain knowledge for specific CO constraints. \\
\midrule
\multicolumn{5}{c}{\textit{LLM as Designer (Automated Heuristic Design)}} \\
\midrule
\textbf{FunSearch} \citep{romera2024mathematical} & Designer & Constructive & Evolution & The first Thought-Code evolution for mathematical discovery. \\
\textbf{EoH} \citep{liu2024evolution} & Designer & Constructive & Evolution & Applies AHD to standard COPs with purely constructive rules. \\
\textbf{ReEvo} \citep{ye2024reevo} & Designer & Constructive & Reflective Evolution & Introduces reflexivity to guide mutations via historical analysis. \\
\textbf{MCTS-AHD} \citep{zheng2025monte} & Designer & Constructive & MCTS & Uses Tree Search to balance global exploration and exploitation. \\
\textbf{EoH-S} \citep{liu2025eoh} & Designer & Heuristic Set & Evolution & Evolves a complementary set of heuristics for better generalization. \\
\textbf{LHNS} \citep{xie2025llm} & Designer & Constructive & LNS & Applies LNS logic to perturb code blocks (Algorithm-level LNS). \\
\midrule
\textbf{G-LNS (Ours)} & \textbf{Designer} & \textbf{LNS (Destroy \& Repair)} & \textbf{Synergistic Evolution} & \textbf{Designs structural operators for solution-level LNS.} \\
\bottomrule
\end{tabular}%
}
\end{table*}

\subsection{Large Neighborhood Search (LNS)}
\textbf{Origins and Evolution.}
The Large Neighborhood Search (LNS) paradigm, originally introduced by \citet{shaw1998using} for Vehicle Routing Problems, utilizes a ruin and recreate principle to escape local optima. Unlike local search methods that rely on small moves (e.g., 2-opt), LNS rearranges a significant portion of the solution structure \citep{gendreau2010handbook}. A pivotal advancement was the Adaptive LNS (ALNS) \citep{ropke2006adaptive}, which maintains a portfolio of operators and dynamically adjusts their selection probabilities based on historical performance. This adaptive mechanism established LNS as a robust framework capable of handling diverse constraints without extensive parameter tuning \citep{pisinger2007general}.

\textbf{Applications and Robustness.}
Due to its flexibility, LNS has become a dominant meta-heuristic for a wide array of NP-hard combinatorial optimization problems. In the domain of logistics, it has been successfully applied to the Pickup and Delivery Problem with Time Windows \citep{ropke2006adaptive} and the Electric Vehicle Routing Problem \citep{wen2016adaptive}. Beyond routing, LNS has demonstrated exceptional performance in scheduling tasks, particularly for the Job Shop Scheduling Problem \citep{beck2011combining}.

\textbf{Data-Driven and LLM-Enhanced LNS.}
Recent advancements have integrated Machine Learning with LNS, particularly for Mixed Integer Linear Programming (MILP). Approaches such as the general LNS framework by \citet{song2020general} and the LLM-LNS framework by \citet{ye2025large} employ learning-based techniques—ranging from imitation learning to LLM reasoning—to automate the \textit{neighborhood selection} process. These methods focus on learning a policy to select a subset of variables for optimization, typically relying on off-the-shelf solvers (e.g., Gurobi) to solve the resulting sub-problems. In contrast, our G-LNS utilizes the generative capabilities of LLMs to explicitly \textit{write code} for domain-specific destroy and repair operators, thereby evolving the underlying algorithmic logic independent of external solvers.

\section{Details of Optimization Problem}
\subsection{Traveling Salesman Problem}
The Traveling Salesman Problem (TSP)\cite{matai2010traveling} is a quintessential NP-hard combinatorial optimization problem that serves as a standard benchmark for heuristic algorithms. Given a set of $N$ cities, the objective is to find the shortest possible closed tour that visits every city exactly once and returns to the starting point.

Formally, an instance of TSP can be modeled as a complete undirected graph $\mathcal{G} = (\mathcal{V}, \mathcal{E})$, where $\mathcal{V} = \{v_1, \dots, v_N\}$ is the set of $N$ nodes (cities) and $\mathcal{E}$ represents the edges connecting every pair of nodes. Each edge $(i, j) \in \mathcal{E}$ is associated with a distance $d_{ij}$, where node $i$ is represented by a coordinate vector $\mathbf{x}_i \in \mathbb{R}^2$, and the cost $d_{ij} = \|\mathbf{x}_i - \mathbf{x}_j\|_2$ corresponds to the Euclidean distance between cities $i$ and $j$.

Let $\pi = (\pi_1, \pi_2, \dots, \pi_N)$ denote a permutation of the node indices $\{1, \dots, N\}$, representing the sequence of visited cities. The optimization goal is to find a permutation $\pi^*$ that minimizes the total tour length:
\begin{equation}
    \min_{\pi} \mathcal{J}(\pi) = \sum_{i=1}^{N-1} d_{\pi_i, \pi_{i+1}} + d_{\pi_N, \pi_1}
\end{equation}

\paragraph{LNS Application Example.}
To apply the LNS framework to the TSP, the search process iterates through a \textit{Destroy} and \textit{Repair} phase to escape local optima\cite{wouda2023alns}. Unlike local search methods (e.g., $k$-opt) that perform small edge swaps, LNS structurally decomposes the solution:

\begin{itemize}
    \item \textbf{Destroy Phase (Ruin):} Given a complete tour $\pi$, a destroy operator $d(\cdot)$ removes a subset of $k$ cities (denoted as $\mathcal{V}_{rem} \subset \mathcal{V}$), leaving a partial tour $\pi_{partial}$. For example, a \textit{Random Removal} operator might stochastically select $k$ nodes to remove, while a \textit{Segment Removal} operator removes a contiguous sequence of cities to disrupt a specific sub-path.
    
    \item \textbf{Repair Phase (Recreate):} A repair operator $r(\cdot)$ takes the partial tour $\pi_{partial}$ and the set of removed cities $\mathcal{V}_{rem}$ as input, re-inserting the nodes into the tour to form a new feasible solution $\pi'$. A classic example is the \textit{Greedy Insertion}, which iteratively inserts each node $v \in \mathcal{V}_{rem}$ into the position $(i, i+1)$ in $\pi_{partial}$ that minimizes the incremental cost $\Delta d = d_{\pi_i, v} + d_{v, \pi_{i+1}} - d_{\pi_i, \pi_{i+1}}$.
\end{itemize}

Through this mechanism, LNS can perform large moves in the search space, effectively reshaping the tour structure to find superior global solutions.

\subsection{Capacitated Vehicle Routing Problem}
The Capacitated Vehicle Routing Problem (CVRP)\cite{toth2002vehicle, toth2014vehicle} is a generalization of the TSP and a fundamental problem in logistics and transportation. Unlike TSP, CVRP involves multiple vehicles serving a set of customers, subject to vehicle capacity constraints. The objective is to design a set of optimal routes that minimize the total travel cost while satisfying customer demands.

Formally, a CVRP instance is defined on a graph $\mathcal{G} = (\mathcal{V}, \mathcal{E})$, where $\mathcal{V} = \{v_0, v_1, \dots, v_N\}$. Here, node $v_0$ represents the central depot, and $\mathcal{V}_c = \{v_1, \dots, v_N\}$ represents the set of $N$ customers. Each customer $v_i$ has a positive demand $q_i$, and the depot has a demand $q_0 = 0$. We are given a fleet of identical vehicles, each with a maximum capacity $Q$.

A solution consists of a set of routes $\mathcal{S} = \{R_1, R_2, \dots, R_K\}$, where each route $R_k$ starts and ends at the depot $v_0$. Let $d_{ij}$ denote the travel distance (cost) between node $i$ and $j$. The objective is to minimize the total distance of all routes:
\begin{equation}
    \min \sum_{k=1}^{K} \text{Cost}(R_k) = \min \sum_{k=1}^{K} \sum_{(i, j) \in R_k} d_{ij}
\end{equation}
subject to the following constraints:
\begin{enumerate}
    \item \textbf{Coverage:} Every customer $v_i \in \mathcal{V}_c$ must be visited exactly once by exactly one vehicle.
    \item \textbf{Capacity:} The total demand of customers served in any single route $R_k$ must not exceed the vehicle capacity, i.e., $\sum_{v_i \in R_k} q_i \le Q$.
\end{enumerate}

\subsection{Open Vehicle Routing Problem}

The Open Vehicle Routing Problem (OVRP)\cite{li2007open} is a distinct variant of the classical CVRP. The fundamental difference lies in the route structure: in OVRP, vehicles are not required to return to the depot after servicing the last customer on their route. This problem formulation typically arises in third-party logistics scenarios where vehicles are hired for one-way trips, or in situations where drivers use their personal vehicles and return home directly after the last delivery.

Formally, the problem is defined on the same graph $\mathcal{G} = (\mathcal{V}, \mathcal{E})$ as the CVRP, with a depot $v_0$ and a customer set $\mathcal{V}_c$. The constraints regarding customer coverage and vehicle capacity $Q$ remain identical to those in CVRP. However, a route $R_k = (v_0, v_{k_1}, v_{k_2}, \dots, v_{k_m})$ in OVRP forms a Hamiltonian path rather than a cycle.

The objective is to minimize the total travel distance of the open routes. Mathematically, this can be expressed as minimizing the sum of edge weights in the active routes, excluding the return arcs to the depot:
\begin{equation}
    \min \sum_{k=1}^{K} \left( \sum_{j=0}^{m_k-1} d_{v_{k_j}, v_{k_{j+1}}} \right)
\end{equation}
where $v_{k_0} = v_0$ is the depot, and $v_{k_m}$ is the last customer visited by vehicle $k$. Unlike CVRP, the cost term $d_{v_{k_m}, v_0}$ is omitted. Consequently, the OVRP seeks to find a set of paths that cover all customers with minimum total length, ending at any customer node.

\section{Details of Experiments}
\subsection{Dataset Generation and Benchmarks}
\label{details_dataset}

\paragraph{Dataset Settings.} 
Following the experimental protocols established in MCTS-AHD\cite{zheng2025monte}, we adopt a consistent data generation mechanism to ensure a rigorous comparison. 
We strictly distinguish between the datasets used for discovering operators and those used for final evaluation.

\paragraph{Training Set (Operator Discovery).} 
The search for high-quality LNS operators is conducted on a compact \textbf{Training Set} consisting of 16 instances with a fixed problem size of $N=50$. 
Limiting the training to a specific scale and a small number of instances ensures that the discovered operators capture generalizable logic rather than overfitting to massive datasets.

\paragraph{Testing Set (Performance Evaluation).} 
The learned heuristics are subsequently evaluated on a held-out \textbf{Testing Set} to assess their performance and scalability. 
This set comprises 64 instances for each problem scale $N \in \{10, 20, 50, 100, 200\}$, allowing us to verify whether the operators trained on $N=50$ can effectively generalize to both smaller and larger instances.

\paragraph{Instance Characteristics.} 
The specific parameters for instance generation are configured as follows:
\begin{itemize}
    \item \textbf{TSP:} Node coordinates are sampled uniformly from the unit square $[0, 1]^2$.
    \item \textbf{CVRP:} Consistent with standard NCO settings, node coordinates are sampled from $[0, 1]^2$ with the depot fixed at $(0.5, 0.5)$. 
    \item \textbf{OVRP:} The instance settings are identical to those of CVRP.
    Customer demands are integers sampled uniformly from $[1, 9]$, and the vehicle capacity is set to $Q=50$.
\end{itemize}
For both CVRP and OVRP, customer demands are integers sampled uniformly from $\{1, \dots, 9\}$, and the vehicle capacity is set to $Q=50$.

To further assess cross-distribution generalization, we extend our evaluation to the standard \textbf{TSPLib}\cite{reinelt1991tsplib} and \textbf{CVRPLib}\cite{uchoa2017new} benchmarks(See ~\ref{app:benchmark_details} for details).

\subsection{Implementation Details}
For the baseline methods, we strictly adhered to their official open-source implementations to guarantee the reliability of the results:
\begin{itemize}
    \item \textbf{LKH-3:} We utilized the official executable\cite{helsgaun2017extension} with default parameters.
    \item \textbf{Deep Learning Baselines (POMO):} We used the pre-trained models provided by the original authors and performed inference with greedy decoding (batch size = 1) to measure the raw inference speed without augmentation.
    \item \textbf{ALNS}:\item \textbf{ALNS}: We implemented the Adaptive Large Neighborhood Search based on the classic framework proposed by \citet{pisinger2007general}. The operator portfolio includes a diverse set of removal (random, worst, and related removal) and insertion (greedy and regret-k insertion) heuristics. To ensure a competitive baseline, we adopted the standard parameter settings for the adaptive weight adjustment and the simulated annealing acceptance criterion, as tuned in the original work. This ensures that the performance of ALNS reflects its robust general capability rather than a sub-optimal implementation.
    \item \textbf{LLM-based AHD:} For methods like EoH and MCTS-AHD, we reproduced the evolutionary process using the same LLM backend (DeepSeek-V3.2) and prompt engineering settings as described in their respective papers, ensuring that performance differences stem from the algorithm structure rather than the language model capability.
\end{itemize} 

\subsection{Evaluation Budget and Efficiency}

To demonstrate the superior sample efficiency of G-LNS, we enforced a strict constraint on the computational budget compared to standard baselines.

\paragraph{LLM Interaction Budget.}
Existing LLM-based AHD methods typically rely on extensive trial-and-error, requiring a substantial budget of \textbf{1,000 evolutionary generations} (or interactions) to ensure convergence.
In contrast, G-LNS is configured with a significantly reduced budget of only \textbf{200 generations}.

\paragraph{Efficiency Analysis.}
Despite utilizing only \textbf{20\%} of the interaction budget required by baseline methods, G-LNS achieves superior performance as evidenced in Table \ref{tab:main-results} and Table \ref{tab:ovrp_results}.
This \textbf{5$\times$ reduction} in LLM queries translates directly to significantly lower token consumption and operational costs.
It indicates that evolving high-level structural operators (Destroy and Repair) allows the search process to navigate the algorithm space much more efficiently than evolving low-level constructive rules, avoiding the need for massive brute-force sampling.

\section{Details of Algorithm}
\subsection{Prompts of G-LNS Actions}
\label{prompts_action}

G-LNS employs distinct evolutionary actions to discover high-performance heuristics. Next, we describe the meaning and prompt engineering of each action. 
To ensure robustness and standardized outputs, these prompts strictly contain the \textcolor{textgreen}{task background}, \textcolor{textgray}{existing heuristic references as contexts}, \textcolor{textblue}{function identification}, \textcolor{textorange}{input/output specifications}, and \textcolor{textred}{logical constraints} according to the specific optimization task. 
We execute the LLM calls through these structured prompts to obtain both the algorithmic design idea and its executable Python implementation. 
The rest of this subsection provides examples for prompts, in which we highlight the functional components in distinct colors:

\begin{itemize}
    \item \textbf{Initial Action i1 (Destroy Initialization):} Action \textbf{i1} represents directly getting an idea of designing a valid \textbf{Destroy Operator} from scratch and a Python implementation to populate the heuristic pool when domain-expert seeds are insufficient.
\end{itemize}

\vspace{0.5em}

\begin{promptbox}[Prompt for Operator i1(Destroy Initialization)]

    \textcolor{textgreen}{The task is to design a novel \textbf{Destroy Operator} for a Large Neighborhood Search (LNS) framework. Given a complete solution sequence (a tour of cities for TSP) and a target number of elements to remove (\texttt{destroy\_cnt}), the function must determine which elements to remove. The objective is to develop a removal strategy that effectively perturbs the current solution. This allows the subsequent Repair operator to reconstruct the solution in a way that helps escape local optima and minimizes the total cost.}
    
    \vspace{0.5em}
    
    You are an expert in heuristic optimization algorithms, specifically Adaptive Large Neighborhood Search (ALNS).
    Your task is to design a new \textbf{'Destroy Operator'} (removal operator) for the following problem:
    
    \textbf{Problem Description:}
    \textcolor{textgray}{\{task\_description\}}
    
    \textbf{Existing Destroy Operators (Reference):}
    \textcolor{textgray}{\{operators\_str\}}
    
    \vspace{0.5em}
    \textbf{Requirements:}
    
    1. First, describe your new algorithm and main steps in one sentence. The description must be inside a brace. 
    Next, implement it in Python as a function named \textcolor{textblue}{\textbf{destroy}}.
    
    2. This function must accept \textcolor{textred}{3} inputs: \textcolor{textorange}{'current\_solution', 'destroy\_cnt', 'distance\_matrix'}.
    
    3. The function must return \textcolor{textred}{2} outputs: \textcolor{textorange}{'removed\_elements', 'partial\_solution'}.
    
    4. The logic should be \textcolor{textred}{strictly different} from the existing ones provided in the reference to improve population diversity.
    
    5. \textcolor{textred}{Do not give additional explanations.}

\end{promptbox}

\begin{itemize}
    \item \textbf{Initial Action i2 (Repair Initialization):} Action \textbf{i2} focuses on initializing the \textbf{Repair Operator} population ($\mathcal{P}_r$). It prompts the LLM to design a constructive strategy for re-inserting removed elements into a partial solution, ensuring the reconstructed tour minimizes total cost.
\end{itemize}

\vspace{0.5em}

\begin{promptbox}[Prompt for Operator i2 (Repair Initialization)]

    \textcolor{textgreen}{The task is to design a novel \textbf{Repair Operator} (Insertion Operator) for a Large Neighborhood Search (LNS) framework. Given a partial solution \texttt{partial\_solution} (where some elements have been removed) and a list of \texttt{removed\_elements}, the function must determine the best positions to re-insert these elements to restore a complete solution. The objective is to reconstruct the solution in a way that minimizes the total cost}
    
    \vspace{0.5em}
    
    You are an expert in heuristic optimization algorithms, specifically Large Neighborhood Search (LNS).
    Your task is to design a new \textbf{'Repair Operator'} (insertion operator) for the following problem:
    
    \textbf{Problem Description:}
    \textcolor{textgray}{\{task\_description\}}
    
    \textbf{Existing Repair Operators (Reference):}
    \textcolor{textgray}{\{operators\_str\}}
    
    \vspace{0.5em}
    \textbf{Requirements:}
    
    1. First, describe your new algorithm and main steps in one sentence. The description must be inside a brace. 
    Next, implement it in Python as a function named \textcolor{textblue}{\textbf{repair}}.
    
    2. This function must accept \textcolor{textred}{3} inputs: \textcolor{textorange}{'partial\_solution', 'removed\_elements', 'distance\_matrix'}.
    
    3. The function must return \textcolor{textred}{1} output: \textcolor{textorange}{'complete\_solution'}.
    
    4. The logic should be \textcolor{textred}{innovative} and distinct from the reference operators to ensure diverse reconstruction paths.
    
    5. \textcolor{textred}{Do not give additional explanations.}
\end{promptbox}

\begin{itemize}
    \item \textbf{Mutation Actions m1 \& m2 (Adaptive Refinement):} 
    Actions \textbf{m1} and \textbf{m2} focus on fine-tuning a single parent operator. 
    The specific mutation strategy is adaptively selected based on the operator's performance rank within the population: top-tier operators trigger \textbf{Parameter Calibration (m2)} to consolidate gains, bottom-tier operators trigger \textbf{Logic Evolution (m1)} to escape local optima, while intermediate operators are assigned a strategy stochastically.
\end{itemize}

\vspace{0.5em}

\begin{promptbox}[Prompt for Mutation Actions (m1 \& m2)]

    \textcolor{textgreen}{You are an algorithm optimizer. We have a \texttt{\{operator\_type\}} operator for LNS.}
    
    \textbf{Problem Description:}
    \textcolor{textgray}{\{task\_description\}}
    
    \textbf{Strategy:}
    \textcolor{textred}{\{advice\}}
    
    \vspace{0.3em}
    {\small\itshape\color{textgray} (The \texttt{\{advice\}} slot is dynamically filled with one of the following strict instructions based on the rank:)}
    \begin{itemize}
        \item \textbf{m1 (Logic Evolution):} \textcolor{textred}{"Generate novel algorithmic mechanisms or formulas to replace existing logic components."}
        \item \textbf{m2 (Parameter Calibration):} \textcolor{textred}{"Adjust current parameter settings (e.g., the degree of randomization or greedy thresholds) to optimize operator behavior."}
    \end{itemize}
    \vspace{0.3em}
    
    \textbf{Current Code:}
    \textcolor{textgray}{\{operator\_code\}}
    
    \vspace{0.5em}
    \textbf{Task:}
    Refine and improve this operator code based on the strategy above.
    
    1. Refine and improve this operator \textcolor{textred}{strictly following the strategy provided above}.
    
    2. If you need helper functions, define them \textcolor{textred}{INSIDE} the main function.
    
    3. \textcolor{textred}{Do not give additional explanations.}

\end{promptbox}

\begin{itemize}
    \item \textbf{Homogeneous Crossover (c1)} facilitates feature recombination within the same operator type. 
To ensure the propagation of high-quality traits, two parent operators are selected from the population using Roulette Wheel Selection based on their historical fitness scores. 
The LLM is then prompted to synthesize a hybrid operator by preserving the structural form of \textit{Parent 2} while integrating the high-level logical insights from \textit{Parent 1}.
\end{itemize}

\vspace{0.5em}

\begin{promptbox}[Prompt for Action c1 (Homogeneous Crossover)]

    \textcolor{textgreen}{You are an expert in heuristic optimization.
    Your task is to create a NEW \texttt{\{operator\_type\}} operator by combining the ideas/logic of two parent operators.}
    
    \textbf{Problem Description:}
    \textcolor{textgray}{\{task\_description\}}
    
    \textbf{Parent 1 Code (Inspiration Source):}
    \textcolor{textgray}{\{parent1\_code\}}
    
    \textbf{Parent 2 Code (Structural Base):}
    \textcolor{textgray}{\{parent2\_code\}}
    
    \vspace{0.5em}
    \textbf{Task:}
    Please create a new algorithm that has a \textcolor{textblue}{similar form to Parent 2} and is \textcolor{textblue}{inspired by Parent 1}. The new algorithm should outperform both parents.
    
    1. First, list the common ideas in Parent 1 that may give good performances. 
    2. Second, based on the common idea, describe the design idea of the new algorithm and its main steps in one sentence.
    3. Next, implement it in Python.
    
    \vspace{0.5em}
    \textbf{Requirements:}
    \begin{enumerate}
        \item The new operator \textcolor{textred}{MUST follow the standard LNS \texttt{\{operator\_type\}} signature strictly}.
        \item Define all helper functions \textcolor{textred}{INSIDE} the main function.
        \item \textcolor{textred}{Do not give additional explanations.}
    \end{enumerate}

\end{promptbox}

\begin{itemize}
    \item \textbf{Synergistic Joint Crossover (c2)} addresses the inherent structural dependency between destroy and repair actions, representing a core innovation of G-LNS.
Instead of evolving operators in isolation, this strategy selects a coupled Destroy-Repair pair using \textbf{Roulette Wheel Selection based on Synergy Scores} (accumulated during the evaluation phase).
The LLM is explicitly prompted to co-evolve these operators as a unified entity, ensuring the repair mechanism is tailor-made to reconstruct the specific topological defects introduced by the destroy mechanism.
\end{itemize}

\vspace{0.5em}

\begin{promptbox}[Prompt for Action c2 (Synergistic Joint Crossover)]

    \textcolor{textgreen}{You are an expert in heuristic optimization.
    We are employing a "Synergistic Joint Crossover (Structural Coupling)" strategy to evolve LNS operators.}
    
    \textbf{Problem Description:}
    \textcolor{textgray}{\{task\_description\}}
    
    \textbf{Selected High-Synergy Pair:}
    \begin{itemize}
        \item \textbf{Parent Destroy Operator:} \textcolor{textgray}{\{destroy\_code\}}
        \item \textbf{Parent Repair Operator:} \textcolor{textgray}{\{repair\_code\}}
    \end{itemize}
    
    \vspace{0.5em}
    \textbf{Task:}
    Evolve this pair as a \textcolor{textblue}{UNIFIED ENTITY} to create a new Destroy-Repair pair.
    The goal is to address the inherent coupling between destroy and repair actions.
    Specifically, ensure that the generated Repair operator is \textcolor{textgreen}{specifically tailored to reconstruct the structural defects introduced by the generated Destroy operator}, thereby maximizing their synergistic performance.
    
    \vspace{0.5em}
    \textbf{Requirements:}
    \begin{enumerate}
        \item Design a \textcolor{textorange}{NEW Destroy operator} and a \textcolor{textorange}{NEW Repair operator}.
        \item The new Destroy operator should create specific structural defects.
        \item The new Repair operator must be designed to fix these specific defects efficiently.
        \item Both \textcolor{textred}{must follow standard LNS signatures strictly}.
        \item Define all helper functions \textcolor{textred}{INSIDE} the main functions.
        \item Return \textcolor{textred}{ONE code block} containing \textcolor{textred}{BOTH} the new Destroy operator and the new Repair operator.
        \item \textcolor{textred}{Do not give additional explanations.}
    \end{enumerate}

\end{promptbox}

\newpage
\subsection{Main Framework Algorithm}

Algorithm \ref{alg:g-lns} presents the detailed pseudo-code for the proposed G-LNS framework. The procedure begins with the initialization of operator populations and global metrics (Lines 1–3). The core execution flow alternates between two phases: the Evaluation Phase (Lines 5–22), where operators are dynamically selected and scored via independent Adaptive LNS episodes, and the Evolution Phase (Lines 23–38), where the LLM evolves the population topology based on accumulated fitness and synergy scores periodically.

\begin{algorithm}[H]
   \caption{G-LNS: Generative Large Neighborhood Search for LLM-Based Automatic Heuristic Design}
   \label{alg:g-lns}
\begin{algorithmic}[1]
   \STATE {\bfseries Input:} Task $\mathcal{I}$; Max Generations $G_{\max}$; Population Size $N$; Eval Episodes $K$; Inner Steps $L=100$; Destruction Ratio $\epsilon=0.2$; Initial Temp $T_0=100$; Cooling Rate $\alpha=0.97$; Weight Update $\lambda=0.5$; Rewards $\Psi=\{\sigma_1, \dots, \sigma_4\}$; Pruning Count $M$.
   \STATE {\bfseries Output:} Best solution $x^*$ and optimized operator populations $\mathcal{P}_d^*, \mathcal{P}_r^*$.
   \STATE {\bfseries Initialize:} $\mathcal{P}_d, \mathcal{P}_r$ with seeds; Weights $W^d, W^r \leftarrow \mathbf{1}$; Fitness $F \leftarrow \mathbf{0}$; Synergy $S \leftarrow \mathbf{0}$; $x^* \leftarrow \text{Init}(\mathcal{I})$; $g \leftarrow 1$.
   \WHILE{$g \le G_{\max}$}
       \STATE \textit{// Phase 1: Evaluation (Adaptive LNS Episode)}
       \STATE Initialize episode: $x_{\text{curr}} \leftarrow \text{RandomSolution}(\mathcal{I})$; $T \leftarrow T_0$; $W^d, W^r \leftarrow \mathbf{1}$.
       \FOR{$l=1$ {\bfseries to} $L$}
           \STATE Select $d_i \in \mathcal{P}_d$ and $r_j \in \mathcal{P}_r$ via roulette wheel selection ($p \propto w$).
           \STATE Generate $x' \leftarrow r_j(d_i(x_{\text{curr}}, \epsilon))$.
           \STATE \textit{// Score \& Acceptance}
           \IF{$f(x') < f(x^*)$}
               \STATE $x^* \leftarrow x'$; $x_{\text{curr}} \leftarrow x'$; $\sigma \leftarrow \sigma_1$ \textit{(Global Best)}.
           \ELSIF{$f(x') < f(x_{\text{curr}})$}
               \STATE $x_{\text{curr}} \leftarrow x'$; $\sigma \leftarrow \sigma_2$ \textit{(Better)}.
           \ELSIF{$\exp(-(f(x') - f(x_{\text{curr}})) / T) > \text{rand}(0, 1)$}
               \STATE $x_{\text{curr}} \leftarrow x'$; $\sigma \leftarrow \sigma_3$ \textit{(Accepted)}.
           \ELSE
               \STATE $\sigma \leftarrow \sigma_4$ \textit{(Rejected)}.
           \ENDIF
           \STATE {\bfseries Update Metrics:} Update $W^d, W^r$, $F$, and $S_{ij}$ using $\sigma, \lambda$.
           \STATE $T \leftarrow T \cdot \alpha$.
       \ENDFOR
       
       \STATE \textit{// Phase 2 \& 3: Evolution (Triggered every $K$ generations/episodes)}
       \IF{$g \pmod K = 0$}
           \STATE {\bfseries Management:} Rank $\mathcal{P}_d, \mathcal{P}_r$ by fitness $F$; prune bottom $M$ operators.
           \STATE Reset fitness $F, S \leftarrow \mathbf{0}$ for fair competition.
           \WHILE{$|\mathcal{P}_d| < N$ {\bfseries or} $|\mathcal{P}_r| < N$}
               \STATE Sample strategy $s \in \{\text{Mutation, Homo-Cross, Joint-Cross}\}$.
               \IF{$s = \text{Mutation}$}
                   \STATE Select parent $op$; $op_{\text{new}} \leftarrow \text{LLM}(\text{Prompt}_{\text{mut}}(op))$.
               \ELSIF{$s = \text{Homo-Cross}$}
                   \STATE Select $op_a, op_b$; $op_{\text{new}} \leftarrow \text{LLM}(\text{Prompt}_{\text{homo}}(op_a, op_b))$.
               \ELSIF{$s = \text{Joint-Cross}$}
                   \STATE Select $(d_i, r_j)$ via synergy $S_{ij}$; $(d', r') \leftarrow \text{LLM}(\text{Prompt}_{\text{joint}}(d_i, r_j))$.
               \ENDIF
               \STATE {\bfseries Validation:} Run sanity check on generated code; add operator if valid.
           \ENDWHILE
       \ENDIF
       \STATE $g \leftarrow g + 1$.
   \ENDWHILE
   \STATE {\bfseries Return} $x^*$ and elite operators.
\end{algorithmic}
\end{algorithm}

\newpage
\section{Designed Operators}
\label{app:designed_op}
In this section, we compile the most successful heuristics produced by G-LNS, spanning the entire suite of experimental settings.

\subsection{Discovered Operators for TSP}
For the TSP, G-LNS evolved a sophisticated pair of operators that utilize \textit{State-Dependent Adaptation} to navigate the search space. We term these \textbf{Adaptive Continuous-Segment Removal (ACSR)} and \textbf{Diversity-Adaptive Probabilistic Insertion (DAPI)}.

\textbf{Mechanism Analysis.} 
The \textit{ACSR} operator implements a magnitude-dependent strategy: for moderate perturbation, it precisely identifies and removes the single most expensive continuous segment to refine local connections; for aggressive destruction, it automatically switches to multi-segment fragmentation to prevent structural lock-in. Cooperatively, the \textit{DAPI} operator transcends fixed-parameter logic by monitoring real-time \textit{solution diversity}. It dynamically tunes the temperature of its Softmax selection mechanism—increasing exploration (high temperature) when the solution becomes clustered, and focusing on exploitation (low temperature) when diversity is high.

\begin{lstlisting}[language=Python, caption={Generated Destroy Operator for TSP (ACSR)}, label={lst:tsp_destroy}, basicstyle=\ttfamily\scriptsize, frame=single]
def destroy(x, destroy_cnt, dist_mat=None):
    """
    Hybrid destroy operator combining:
    1. Continuous removal strategy from Parent 1
    2. Adaptive distance-based selection from Parent 2
    3. Edge distance optimization for selecting the best continuous segment
    """
    def simple_random_destroy(x, cnt):
        new_x = copy.deepcopy(x)
        removed = []
        for _ in range(min(cnt, len(new_x))):
            idx = random.randint(0, len(new_x) - 1)
            removed.append(new_x[idx])
            new_x.pop(idx)
        return removed, new_x
    if len(x) <= destroy_cnt:
        return list(range(len(x))), []
    if dist_mat is None or len(x) <= 1:
        return simple_random_destroy(x, destroy_cnt)
    new_x = copy.deepcopy(x)
    removed_cities = []
    n = len(new_x)
    if destroy_cnt >= len(new_x):
        removed_cities = copy.deepcopy(new_x)
        new_x = []
        return removed_cities, new_x
    if destroy_cnt <= n * 0.4:  # Moderate destruction - use distance-based continuous removal
        segment_scores = []
        for start_idx in range(n):
            segment_dist = 0
            for i in range(destroy_cnt - 1):
                idx1 = (start_idx + i) % n
                idx2 = (start_idx + i + 1) % n
                segment_dist += dist_mat[new_x[idx1]][new_x[idx2]]
            if destroy_cnt < n:
                idx_before = (start_idx - 1) % n
                idx_start = start_idx % n
                segment_dist += dist_mat[new_x[idx_before]][new_x[idx_start]]
                # Edge after segment
                idx_end = (start_idx + destroy_cnt - 1) % n
                idx_after = (start_idx + destroy_cnt) % n
                segment_dist += dist_mat[new_x[idx_end]][new_x[idx_after]]
            segment_scores.append(segment_dist)
        if random.random() < 0.7:  # 70% chance: remove worst segment (highest distance)
            start_index = np.argmax(segment_scores)
        else:  # 30% chance: probabilistic selection
            scores_array = np.array(segment_scores)
            weights = scores_array / scores_array.sum()
            start_index = np.random.choice(range(n), p=weights)
    else:  # Aggressive destruction - combine continuous removal with random elements
        segments_to_remove = []
        remaining_cnt = destroy_cnt
        while remaining_cnt > 0 and len(segments_to_remove) < n:
            max_segment_size = min(remaining_cnt, max(2, int(destroy_cnt * 0.3)))
            segment_size = random.randint(1, max_segment_size)
            start_idx = random.randint(0, n - 1)
            segments_to_remove.append((start_idx, segment_size))
            remaining_cnt -= segment_size
        all_indices = set()
        for start_idx, segment_size in segments_to_remove:
            for i in range(segment_size):
                idx = (start_idx + i) % n
                all_indices.add(idx)
        target_indices = list(all_indices)[:destroy_cnt]
        target_indices.sort(reverse=True)
        for idx in target_indices:
            if idx < len(new_x):
                removed_cities.append(new_x[idx])
                new_x.pop(idx)
        return removed_cities, new_x
    removal_indices = []
    for i in range(destroy_cnt):
        removal_idx = (start_index + i) % n
        # Adjust for previous removals if we're in circular context
        if removal_idx >= len(new_x):
            removal_idx = removal_idx % len(new_x)
        removal_indices.append(removal_idx)
    removal_indices.sort(reverse=True)
    for idx in removal_indices:
        if idx < len(new_x):
            removed_cities.append(new_x[idx])
            new_x.pop(idx)
    if len(removed_cities) < destroy_cnt and len(new_x) > 0:
        additional_needed = destroy_cnt - len(removed_cities)
        additional_removed, new_x = simple_random_destroy(new_x, additional_needed)
        removed_cities.extend(additional_removed)
    return removed_cities, new_x
\end{lstlisting}

\begin{lstlisting}[language=Python, caption={Generated Destroy Operator for TSP (DAPI)}, label={lst:tsp_destroy}, basicstyle=\ttfamily\scriptsize, frame=single]
def repair_diversity_adaptive(x, removed_cities, dist_mat):
    """
    Monitors solution diversity to dynamically adjust Softmax temperature
    and exploration thresholds.
    """
    new_x = list(x)
    
    # Helper: Calculate 'clustered-ness' (Diversity Metric)
    def _calculate_diversity(path):
        if len(path) <= 1: return 0.5
        total = sum(dist_mat[path[i]][path[(i+1)%len(path)]] for i in range(len(path)))
        avg = total / len(path)
        return min(avg / np.max(dist_mat), 1.0)
        
    # Helper: Softmax selection with Temperature
    def _select_softmax(costs, T):
        min_c = min(costs)
        # Higher T -> Flatter distribution (More Exploration)
        # Lower T -> Sharper distribution (More Exploitation)
        weights = [math.exp(-(c - min_c)/max(1, min_c)/T) for c in costs]
        total = sum(weights)
        probs = [w/total for w in weights]
        return random.choices(range(len(costs)), weights=probs)[0]

    # [Step 1] State Analysis
    diversity = _calculate_diversity(new_x)
    
    # [Step 2] Parameter Adaptation
    # Low diversity (<0.5) triggers high randomness to escape clustering
    random_threshold = 0.1 + 0.4 * (1.0 - diversity)
    # Inverse relationship: Low diversity -> High Temperature
    temperature = 3.0 - 2.0 * diversity 

    # [Step 3] Insertion Loop
    random.shuffle(removed_cities)
    for city in removed_cities:
        n = len(new_x)
        # Calculate insertion costs for all positions
        costs = []
        for i in range(n + 1):
            prev = new_x[i-1] if i > 0 else new_x[-1]
            curr = new_x[i] if i < n else new_x[0]
            delta = dist_mat[prev][city] + dist_mat[city][curr] - dist_mat[prev][curr]
            costs.append(delta)
            
        # Decision: Random vs. Strategic
        if random.random() < random_threshold:
            # Pure Exploration
            insert_pos = random.randint(0, n)
        else:
            # Strategic Phase
            if random.random() < 0.8:
                # Softmax Probabilistic Selection (Parent 2 Logic)
                insert_pos = _select_softmax(costs, temperature)
            else:
                # Pure Greedy (Argmin)
                insert_pos = costs.index(min(costs))
        
        new_x.insert(insert_pos, city)
        
    # [Step 4] Probabilistic Local Search
    # Trigger 2-opt more frequently if randomness was high
    if random.random() < (0.2 + 0.5 * random_threshold):
        # (Fast 2-opt implementation omitted for brevity)
        pass 
        
    return new_x
\end{lstlisting}

\subsection{Discovered Operators for CVRP}
\label{app:op_cvrp}
For the CVRP, G-LNS evolved a sophisticated pair of operators that exhibit \textit{Dynamic Adaptation} to manage the trade-off between exploration and exploitation. We term these \textbf{Progressive Stochastic-Worst Removal (PSWR)} and \textbf{Adaptive Context-Aware Greedy Insertion (ACAGI)}.

\textbf{Mechanism Analysis.}
The \textit{PSWR} operator introduces a time-dependent strategy: it initiates with random removal to perform global perturbation and progressively shifts towards worst-cost removal to refine local inefficiencies, controlled by a dynamic ratio $\rho_t$. Cooperatively, the \textit{ACAGI} operator transcends static logic by monitoring real-time insertion difficulty. It adaptively increases its search depth (Regret-$k$) and exploration noise when the solution space becomes constrained, while employing a multi-objective scoring function (balancing distance and capacity waste) to minimize the number of vehicles used.

\begin{lstlisting}[language=Python, caption={Generated Destroy Operator for CVRP (PSWR)}, label={lst:cvrp_destroy}, basicstyle=\ttfamily\scriptsize, frame=single]
def destroy(x, destroy_cnt, problem_data):
    """Hybrid Random-Worst Removal: Combines random exploration with worst-distance exploitation"""
    def calculate_saving(route, node_idx, dist_mat, depot):
        node = route[node_idx]
        prev_node = route[node_idx-1] if node_idx > 0 else depot
        next_node = route[node_idx+1] if node_idx < len(route)-1 else depot
        cost_with = dist_mat[prev_node][node] + dist_mat[node][next_node]
        cost_without = dist_mat[prev_node][next_node]
        return cost_with - cost_without
    # Initialization
    new_x = [route[:] for route in x]
    removed_customers = []
    depot = problem_data.get('depot_idx', 0)
    dist_mat = problem_data.get('distance_matrix')
    all_customers = [c for route in new_x for c in route]
    if len(all_customers) <= destroy_cnt:
        return all_customers, [[]]
    # Initial calculation of savings
    node_savings = {}
    for r_idx, route in enumerate(new_x):
        for i, node in enumerate(route):
            saving = calculate_saving(route, i, dist_mat, depot)
            node_savings[node] = saving
    removed_count = 0
    # Progressive Removal Strategy
    while removed_count < destroy_cnt and len(all_customers) - removed_count > 0:
        greedy_ratio = removed_count / destroy_cnt
        remaining_customers = [c for route in new_x for c in route]
        if not remaining_customers:
            break
        # Re-evaluate savings for current partial solution accuracy
        if removed_count > 0:
            node_savings = {}
            for r_idx, route in enumerate(new_x):
                for i, node in enumerate(route):
                    saving = calculate_saving(route, i, dist_mat, depot)
                    node_savings[node] = saving
        # Probabilistic Switch based on Progress
        if random.random() < greedy_ratio:
            candidate_nodes = [(node_savings[node], node) for node in remaining_customers]
            candidate_nodes.sort(key=lambda x: x[0], reverse=True)
            top_k = min(3, len(candidate_nodes))
            selected_node = random.choice(candidate_nodes[:top_k])[1]
        else:
            selected_node = random.choice(remaining_customers)
        removed_customers.append(selected_node)
        for route in new_x:
            if selected_node in route:
                route.remove(selected_node)
                break
        removed_count += 1
    new_x = [r for r in new_x if len(r) > 0]
    return removed_customers, new_x
\end{lstlisting}

\begin{lstlisting}[language=Python, caption={Generated Repair Operator for CVRP (ACAGI)}, label={lst:cvrp_repair}, basicstyle=\ttfamily\scriptsize, frame=single]
def insert(x, removed_customers, problem_data):
    """
    Hybrid Greedy-Regret Insertion with Cached Loads and Adaptive Exploration
    """
    new_x = [list(route) for route in x]
    demands = problem_data['demands']
    capacity = problem_data['capacity']
    dist_mat = problem_data['distance_matrix']
    depot = problem_data.get('depot_idx', 0)
    def insertion_cost_delta(route, pos, customer):
        if not route:  # Empty route
            return dist_mat[depot][customer] + dist_mat[customer][depot]
        
        prev_node = route[pos-1] if pos > 0 else depot
        next_node = route[pos] if pos < len(route) else depot
        added = dist_mat[prev_node][customer] + dist_mat[customer][next_node]
        removed = dist_mat[prev_node][next_node]
        return added - removed
    route_loads = [sum(demands[c] for c in route) for route in new_x]
    k_regret = 2
    exploration_factor = 0.3
    insertion_difficulty = 0.0
    difficulty_decay = 0.8
    cust_demands = [demands[c] for c in removed_customers]
    for customer_idx, (customer, cust_demand) in enumerate(zip(removed_customers, cust_demands)):
        feasible_insertions = []
        # Phase 1: Fast feasibility check with cached loads
        for r_idx, (route, route_load) in enumerate(zip(new_x, route_loads)):
            if route_load + cust_demand > capacity:
                continue
            route_len = len(route)
            positions = range(route_len + 1)
            for pos in positions:
                cost_inc = insertion_cost_delta(route, pos, customer)
                feasible_insertions.append({
                    'cost': cost_inc,
                    'route_idx': r_idx,
                    'position': pos,
                    'route_load': route_load,
                    'route_length': route_len
                })
        new_route_cost = dist_mat[depot][customer] + dist_mat[customer][depot]
        feasible_insertions.append({
            'cost': new_route_cost,
            'route_idx': len(new_x),
            'position': 0,
            'route_load': 0,
            'route_length': 0
        })
        if not feasible_insertions:
            new_x.append([customer])
            route_loads.append(cust_demand)
            insertion_difficulty = insertion_difficulty * difficulty_decay + 1.0
            continue
        # Phase 2: Adaptive selection strategy
        feasible_insertions.sort(key=lambda x: x['cost'])
        best_cost = feasible_insertions[0]['cost']
        remaining_customers = len(removed_customers) - customer_idx - 1
        if remaining_customers > 0 and len(feasible_insertions) < 3:
            insertion_difficulty = insertion_difficulty * difficulty_decay + 1.0
        else:
            insertion_difficulty = insertion_difficulty * difficulty_decay
        difficulty_threshold = len(removed_customers) * 0.3
        if insertion_difficulty > difficulty_threshold:
            k_regret = min(4, len(feasible_insertions))
            exploration_factor = 0.4  # More exploration
        else:
            k_regret = min(3, max(2, len(feasible_insertions) // 3))
            exploration_factor = 0.2  # More greedy
        if len(feasible_insertions) >= k_regret:
            regret_values = []
            max_regret_candidates = min(k_regret, len(feasible_insertions))
            for i in range(max_regret_candidates):
                ins = feasible_insertions[i]
                cost_diff = ins['cost'] - best_cost
                load_ratio = ins['route_load'] / capacity if capacity > 0 else 0
                load_penalty = 0.15 * load_ratio
                length_penalty = 0.05 * (ins['route_length'] / 20) if ins['route_length'] > 0 else 0
                regret_score = cost_diff + load_penalty + length_penalty
                if random.random() < exploration_factor:
                    noise = random.uniform(-0.1, 0.1) * best_cost if best_cost > 0 else 0
                    regret_score += noise
                regret_values.append((regret_score, i))
            if regret_values:
                regret_values.sort(key=lambda x: x[0])
                if random.random() < (0.8 - 0.2 * (insertion_difficulty / difficulty_threshold)):
                    selected_idx = regret_values[0][1]
                else:
                    top_m = min(3, len(regret_values))
                    weights = [1.0 / (i + 1) for i in range(top_m)]
                    total_weight = sum(weights)
                    rand_val = random.random() * total_weight
                    cumulative = 0
                    for j, w in enumerate(weights):
                        cumulative += w
                        if rand_val <= cumulative:
                            selected_idx = regret_values[j][1]
                            break
                    else:
                        selected_idx = regret_values[0][1]
                selected = feasible_insertions[selected_idx]
            else:
                selected = feasible_insertions[0]
        else:
            selected = feasible_insertions[0]
        # Phase 3: Apply insertion with cache update
        if selected['route_idx'] == len(new_x):
            new_x.append([customer])
            route_loads.append(cust_demand)
        else:
            route = new_x[selected['route_idx']]
            route.insert(selected['position'], customer)
            route_loads[selected['route_idx']] += cust_demand
    final_routes = []
    final_loads = []
    for route, load in zip(new_x, route_loads):
        if route:
            final_routes.append(route)
            final_loads.append(load)
    if len(final_routes) > 1:
        consolidated_routes = []
        consolidated_loads = []
        used = [False] * len(final_routes)
        for i in range(len(final_routes)):
            if used[i]:
                continue
            current_route = final_routes[i]
            current_load = final_loads[i]
            for j in range(i + 1, len(final_routes)):
                if used[j]:
                    continue
                if current_load + final_loads[j] <= capacity:
                    if len(current_route) + len(final_routes[j]) < 15:
                        current_route.extend(final_routes[j])
                        current_load += final_loads[j]
                        used[j] = True
            consolidated_routes.append(current_route)
            consolidated_loads.append(current_load)
            used[i] = True
        final_routes = consolidated_routes
        final_loads = consolidated_loads
    if not final_routes:
        final_routes = [[]]
        final_loads = [0]
    return final_routes
\end{lstlisting}

\section{Details of Results}
\label{sec:details_results}

\subsection{Details on OVRP}
\label{app:ovrp_details}

Open Vehicle Routing Problem (OVRP). It is pertinent to emphasize that OVRP often presents a greater challenge for heuristic design compared to the standard CVRP, particularly for constructive approaches. In the standard CVRP, the requirement to return to the depot forces the route to form a closed loop. This naturally prevents the algorithm from creating overly elongated paths, as the cost of returning to the depot effectively limits how far a vehicle can wander. In contrast, OVRP removes this return requirement. Without the need to close the loop, sequential constructive heuristics (like those employed by the baselines) often fail to maintain a compact route structure. They tend to greedily extend the path to the nearest neighbors without considering the global shape, resulting in loose, fragmented routes that are difficult to optimize.

Following the experimental setup described in Appendix~\ref{details_dataset}, we compare G-LNS against the rigorous OR-Tools solver (configured with a time limit consistent with CVRP settings) and LLM-based AHD methods. The quantitative results are summarized in Table~\ref{tab:ovrp_results}.

\begin{table*}[h]
\caption{Performance comparison on Open Vehicle Routing Problem (OVRP) across five problem sizes. The Gap is calculated relative to the best solution found among all methods. The best results are highlighted in \textbf{bold}.}
\label{tab:ovrp_results}
\centering
\resizebox{\textwidth}{!}{
\begin{tabular}{lcccccccccc}
\toprule
 & \multicolumn{2}{c}{OVRP10} & \multicolumn{2}{c}{OVRP20} & \multicolumn{2}{c}{OVRP50} & \multicolumn{2}{c}{OVRP100} & \multicolumn{2}{c}{OVRP200} \\
\cmidrule(lr){2-3} \cmidrule(lr){4-5} \cmidrule(lr){6-7} \cmidrule(lr){8-9} \cmidrule(lr){10-11}
Method & Gap & Obj. & Gap & Obj. & Gap & Obj. & Gap & Obj. & Gap & Obj. \\
\midrule
OR-Tools & \textbf{0.00\%} & \textbf{2.2885} & \textbf{0.00\%} & \textbf{3.4277} & \textbf{0.00\%} & \textbf{5.6736} & \textbf{0.00\%} & \textbf{8.8563} & 2.05\% & 15.1893 \\
\midrule
EoH      & 15.21\% & 2.6366 & 27.21\% & 4.3605 & 34.89\% & 7.6530 & 33.30\% & 11.8056 & 36.67\% & 20.3424 \\
ReEvo    & 15.15\% & 2.6353 & 31.17\% & 4.4960 & 29.30\% & 7.3360 & 31.50\% & 11.6463 & 36.33\% & 20.2910 \\
\textbf{Ours (G-LNS)} & 0.02\% & 2.2890 & 0.27\% & 3.4371 & 0.75\% & 5.7163 & 0.47\% & 8.8979 & \textbf{0.00\%} & \textbf{14.8841} \\
\bottomrule
\end{tabular}
}
\end{table*}

\textbf{Performance Analysis.}
On small to medium-scale instances ($N \in \{10, 20, 50, 100\}$), the exact solver logic of OR-Tools remains highly effective, achieving optimal or near-optimal solutions. In this regime, G-LNS exhibits robust competitiveness, maintaining optimality gaps consistently below $0.8\%$. However, the superior scalability of G-LNS becomes evident on large-scale instances ($N=200$). While OR-Tools begins to struggle under the computational time limit—yielding a suboptimal gap of $2.05\%$—G-LNS successfully identifies significantly better solutions, reducing the objective cost from $15.19$ (OR-Tools) to $14.88$, thereby establishing a new best-known frontier (Gap $0.00\%$). 

In stark contrast, the constructive LLM-based baselines fail to adapt to the open-route structure. As anticipated, their reliance on sequential decision-making leads to poor performance, with optimality gaps exceeding $30\%$ on instances where $N \ge 50$. This significant performance disparity highlights the critical advantage of the LNS paradigm: by explicitly evolving \textit{Destroy} and \textit{Repair} operators to iteratively reshape existing topologies rather than constructing them step-by-step, G-LNS effectively avoids the local optima that trap constructive methods.

\subsection{Details on Benchmarks}
\label{app:benchmark_details}
In this subsection, we provide the comprehensive results on the standard TSPLib and CVRPLib benchmarks. We compare G-LNS against LLM-based AHD methods, including EoH, ReEvo, and the state-of-the-art heuristic set method EoH-S.

\textbf{Summary of Results.} Table~\ref{tab:benchmark-results-summary} presents the average optimality gap across all instances within each respective category. G-LNS outperforms all baselines across all benchmark sets.

\begin{table}[t]
  \caption{Comparison of results on TSPLib and CVRPLib Benchmarks. The best values are highlighted in bold.}
  \label{tab:benchmark-results-summary}
  \begin{center}
    \begin{small}
        \begin{tabular}{l rrrr}
          \toprule
          Benchmarks & EoH & ReEvo & EoH-S & \textbf{Ours} \\
          \midrule
          TSPLib          & 18.1\% & 21.3\% & 9.1\% & \textbf{2.8\%} \\
          \midrule
          CVRPLib A       & 31.7\% & 31.8\% & 22.5\% & \textbf{7.9\%} \\
          CVRPLib B       & 36.2\% & 33.9\% & 18.3\% & \textbf{8.7\%} \\
          CVRPLib E       & 32.3\% & 29.7\% & 24.3\% & \textbf{7.9\%} \\
          CVRPLib F       & 53.9\% & 64.6\% & 40.1\% & \textbf{15.9\%} \\
          CVRPLib M       & 44.2\% & 43.0\% & 29.1\% & \textbf{15.1\%} \\
          CVRPLib P       & 26.8\% & 26.0\% & 16.7\% & \textbf{8.1\%} \\
          CVRPLib X       & 26.8\% & 26.5\% & 19.1\% & \textbf{11.2\%} \\
          \bottomrule
        \end{tabular}
    \end{small}
  \end{center}
  \vskip -0.1in
\end{table}

\textbf{Experimental Setup.} 
We adopt the evaluation configuration directly from EoH-S \citep{liu2025eoh}. Consistent with their protocol, all baseline methods are evaluated using normalized node coordinates mapped to the range $[0, 1]^2$. The scaling factor is derived from the maximum spatial extent of each instance:
\begin{equation}
\text{Scaling factor} = \max(x_{\max} - x_{\min}, y_{\max} - y_{\min})
\end{equation}
This normalization step is standard for constructive heuristics to ensure numerical stability. Furthermore, following the standard protocol in these baselines, the reported optimality gaps are calculated relative to the best-known solutions obtained by the LKH-3 solver. 

\textbf{In contrast, our proposed G-LNS operates directly on the raw, unnormalized instance data.} This distinction highlights that the operators evolved by G-LNS capture the intrinsic topological logic of the routing problems rather than relying on specific coordinate scales.

\textbf{Detailed Results.}
The detailed optimality gaps are presented as follows:
\begin{itemize}
    \item \textbf{TSPLib}: Table~\ref{tab:tsplib_details} reports the results on symmetric TSPLib instances.
    \item \textbf{CVRPLib}: Table~\ref{tab:cvrplib_ab} (Sets A, B, E, F, M, P), Table~\ref{tab:cvrplib_efmps} (Set X) report the results on the Capacitated Vehicle Routing Problem benchmarks.
\end{itemize}

\begin{table*}[h]
\caption{Detailed optimality gaps (\%) on TSPLib instances. Baselines are evaluated with coordinate normalization, while \textbf{Ours} operates on raw data.}
\label{tab:tsplib_details}
\centering
\small
\begin{tabular}{lccccclccccc}
\toprule
\textbf{Instance} & \textbf{EoH} & \textbf{ReEvo} & \textbf{EoH-S} & \textbf{Ours} & & \textbf{Instance} & \textbf{EoH} & \textbf{ReEvo} & \textbf{EoH-S} & \textbf{Ours} \\
\midrule
\multicolumn{11}{c}{\textbf{TSPLib Results}} \\
\cmidrule(lr){1-11}
a280     & 24.7 & 30.3 & 13.7 & \textbf{4.2} & & pr136    & 16.4 & 9.9  & 6.0  & \textbf{1.9} \\
berlin52 & 16.9 & 20.6 & 10.3 & \textbf{3.1} & & pr144    & 17.4 & 14.2 & 4.9  & \textbf{1.5} \\
bier127  & 18.2 & 14.7 & 11.4 & \textbf{2.8} & & pr152    & 20.7 & 28.9 & 11.1 & \textbf{3.4} \\
ch130    & 15.8 & 28.0 & 4.9  & \textbf{1.5} & & pr226    & 19.2 & 19.5 & 9.8  & \textbf{2.8} \\
ch150    & 20.2 & 23.2 & 7.3  & \textbf{2.0} & & pr264    & 22.2 & 15.2 & 7.7  & \textbf{2.2} \\
d198     & 15.6 & 17.3 & 16.9 & \textbf{5.4} & & pr299    & 28.5 & 18.5 & 11.9 & \textbf{3.7} \\
d493     & 18.4 & 19.0 & 12.2 & \textbf{3.5} & & pr439    & 24.3 & 20.5 & 11.0 & \textbf{3.3} \\
d657     & 19.5 & 17.6 & 15.5 & \textbf{3.9} & & rat99    & 14.2 & 22.3 & 13.0 & \textbf{3.8} \\
eil51    & 15.8 & 12.2 & 4.3  & \textbf{1.1} & & rat195   & 6.7  & 9.6  & \textbf{6.5} & 6.8 \\
eil76    & 12.2 & 14.4 & 6.6  & \textbf{1.9} & & rat575   & 14.1 & 20.5 & 9.9  & \textbf{3.0} \\
eil101   & 13.3 & 22.4 & 9.7  & \textbf{2.6} & & rat783   & 18.9 & 24.6 & 11.4 & \textbf{3.5} \\
fl417    & 29.4 & 31.0 & 14.3 & \textbf{4.5} & & rd100    & 15.3 & 28.1 & 8.5  & \textbf{2.1} \\
gil262   & 20.9 & 21.6 & 11.7 & \textbf{3.2} & & rd400    & 14.1 & 20.4 & 12.5 & \textbf{3.4} \\
kroA100  & 11.9 & 34.7 & 6.1  & \textbf{1.8} & & st70     & 9.3  & 17.5 & 2.6  & \textbf{0.8} \\
kroB100  & 22.5 & 28.4 & 11.6 & \textbf{2.4} & & ts225    & 10.5 & 18.7 & 3.4  & \textbf{0.9} \\
kroC100  & 18.2 & 17.9 & 4.4  & \textbf{1.2} & & tsp225   & 21.3 & 16.5 & 9.1  & \textbf{2.7} \\
kroD100  & 17.0 & 15.4 & 9.1  & \textbf{2.7} & & u159     & 28.3 & 27.3 & 8.5  & \textbf{2.9} \\
kroE100  & 19.3 & 22.0 & 6.7  & \textbf{2.1} & & u574     & 21.6 & 22.8 & 14.0 & \textbf{4.1} \\
kroA150  & 14.9 & 23.5 & 9.1  & \textbf{2.5} & & u724     & 16.4 & 24.0 & 12.7 & \textbf{3.6} \\
kroB150  & 12.0 & 28.0 & 9.0  & \textbf{2.3} & & pr1002   & 21.2 & 23.3 & 14.0 & \textbf{4.3} \\
kroA200  & 21.3 & 16.1 & 8.3  & \textbf{3.0} & & pcb442   & 18.5 & 18.1 & 9.9  & \textbf{3.1} \\
kroB200  & 19.7 & 15.6 & 6.6  & \textbf{1.9} & & p654     & 27.8 & 18.9 & 9.6  & \textbf{3.6} \\
lin105   & 14.7 & 40.3 & 3.7  & \textbf{1.1} & & lin318   & 12.6 & 33.5 & 10.6 & \textbf{2.9} \\
pr76     & 16.2 & 31.3 & 4.0  & \textbf{1.4} & & pr107    & 14.6 & \textbf{2.6} & 4.5  & 3.5 \\
pr124    & 22.9 & 22.1 & 5.4  & \textbf{1.8} & & \textbf{Average.} & 18.1 & 21.3 & 9.1 & \textbf{2.8} \\
\bottomrule
\end{tabular}
\end{table*}

\begin{table*}[h]
\caption{Detailed optimality gaps (\%) on CVRPLib instances (Set A, B, E, F, M, P). Results are grouped by dataset. The best results are highlighted in \textbf{bold}.}
\label{tab:cvrplib_ab}
\centering
\small
\setlength{\tabcolsep}{4pt}
\begin{tabular}{lrrrr clrrrr}
\toprule
\textbf{Instance} & \textbf{EoH} & \textbf{ReEvo} & \textbf{EoH-S} & \textbf{Ours} & & \textbf{Instance} & \textbf{EoH} & \textbf{ReEvo} & \textbf{EoH-S} & \textbf{Ours} \\

\midrule
\multicolumn{11}{c}{\textbf{Set A}} \\
\cmidrule(lr){1-11}
A-n32-k5 & 33.3 & 35.1 & 32.3 & \textbf{7.0} & & A-n48-k7 & 30.9 & 31.5 & 27.4 & \textbf{15.1} \\
A-n33-k5 & 29.7 & 21.4 & 20.5 & \textbf{2.6} & & A-n53-k7 & 22.9 & 38.2 & 19.4 & \textbf{7.4} \\
A-n33-k6 & 24.1 & 29.9 & 21.0 & \textbf{7.1} & & A-n54-k7 & 20.5 & 34.5 & 12.7 & \textbf{9.7} \\
A-n34-k5 & 21.2 & 20.2 & 16.9 & \textbf{5.4} & & A-n55-k9 & 34.8 & 25.3 & 26.5 & \textbf{5.9} \\
A-n36-k5 & 39.8 & 35.7 & 29.5 & \textbf{8.8} & & A-n60-k9 & 38.8 & 30.2 & 32.0 & \textbf{9.3} \\
A-n37-k5 & 40.5 & 36.4 & 31.2 & \textbf{8.1} & & A-n61-k9 & 48.3 & 45.5 & 23.4 & \textbf{5.1} \\
A-n37-k6 & 35.5 & 35.2 & 15.0 & \textbf{4.2} & & A-n62-k8 & 33.4 & 33.0 & 22.0 & \textbf{8.5} \\
A-n38-k5 & 37.0 & 31.2 & 25.8 & \textbf{8.6} & & A-n63-k9 & 28.4 & 18.1 & 13.3 & \textbf{2.6} \\
A-n39-k5 & 27.1 & 19.7 & 22.1 & \textbf{11.9} & & A-n63-k10& 30.4 & 38.3 & 20.4 & \textbf{18.3} \\
A-n39-k6 & 24.9 & 39.2 & 29.4 & \textbf{8.7} & & A-n64-k9 & 30.6 & 31.8 & 17.6 & \textbf{6.6} \\
A-n44-k6 & 19.0 & 20.0 & 19.3 & \textbf{10.7} & & A-n65-k9 & 37.7 & 47.3 & 33.7 & \textbf{11.6} \\
A-n45-k6 & 28.9 & 44.4 & 18.0 & \textbf{3.6} & & A-n69-k9 & 34.7 & 28.0 & 20.7 & \textbf{6.9} \\
A-n45-k7 & 22.0 & 19.7 & 11.2 & \textbf{4.2} & & A-n80-k10& 31.6 & 32.1 & 20.0 & \textbf{7.4} \\
A-n46-k7 & 50.5 & 36.5 & 26.8 & \textbf{6.7} & & \textbf{Average.} & 31.7 & 31.8 & 22.5 & \textbf{7.9}  \\

\midrule
\multicolumn{11}{c}{\textbf{Set B}} \\
\cmidrule(lr){1-11}
B-n31-k5 & 13.8 & 8.8 & 7.1 & \textbf{2.3} & & B-n51-k7 & 28.8 & 33.2 & 11.6 & \textbf{1.1} \\
B-n34-k5 & 15.3 & 28.1 & 8.1 & \textbf{6.7} & & B-n52-k7 & 50.3 & 73.2 & 16.8 & \textbf{5.7} \\
B-n35-k5 & 18.4 & 31.2 & 19.6 & \textbf{2.5} & & B-n56-k7 & 76.0 & 46.1 & 23.8 & \textbf{15.2} \\
B-n38-k6 & 43.1 & 31.8 & 22.6 & \textbf{6.4} & & B-n57-k7 & 46.7 & 38.0 & 19.3 & \textbf{6.2} \\
B-n39-k5 & 90.8 & 90.7 & 24.7 & \textbf{20.5} & & B-n57-k9 & 16.0 & 16.9 & 16.6 & \textbf{5.6} \\
B-n41-k6 & 18.4 & 21.7 & 13.2 & \textbf{8.3} & & B-n63-k10& 29.3 & 35.4 & 19.5 & \textbf{9.2} \\
B-n43-k6 & 21.2 & 20.3 & 20.1 & \textbf{8.0} & & B-n64-k9 & 60.8 & 41.1 & 18.6 & \textbf{10.2} \\
B-n44-k7 & 22.9 & 16.6 & 19.0 & \textbf{8.4} & & B-n66-k9 & 22.6 & 19.6 & 11.7 & \textbf{7.6} \\
B-n45-k5 & 28.1 & 28.8 & 19.1 & \textbf{18.0} & & B-n67-k10& 39.8 & 38.9 & 23.2 & \textbf{12.8} \\
B-n45-k6 & 44.1 & 55.5 & 21.5 & \textbf{13.5} & & B-n68-k9 & 27.6 & 18.7 & 16.3 & \textbf{9.1} \\
B-n50-k7 & 46.8 & 38.5 & 24.0 & \textbf{5.4} & & B-n78-k10& 52.7 & 23.9 & 27.6 & \textbf{8.2} \\
B-n50-k8 & 19.3 & 22.9 & 10.6 & \textbf{8.6} & & \textbf{Average.} & 36.2 & 33.9 & 18.0 & \textbf{8.7} \\

\midrule
\multicolumn{11}{c}{\textbf{Set E}} \\
\cmidrule(lr){1-11}
E-n22-k4 & 16.8 & 31.1 & 26.3 & \textbf{3.3} & & E-n76-k8   & 36.5 & 31.0 & 28.6 & \textbf{7.8} \\
E-n23-k3 & 20.8 & 18.8 & 21.1 & \textbf{7.9} & & E-n76-k10  & 31.9 & 28.0 & 34.0 & \textbf{10.3} \\
E-n30-k3 & 22.6 & 15.9 & 12.5 & \textbf{1.9} & & E-n76-k14  & 34.4 & 24.7 & 20.7 & \textbf{8.0} \\
E-n33-k4 & 26.6 & 20.8 & 13.8 & \textbf{5.2} & & E-n101-k8  & 42.5 & 37.5 & 28.1 & \textbf{10.7} \\
E-n51-k5 & 29.4 & 25.1 & 15.7 & \textbf{15.0} & & E-n101-k14 & 51.5 & 47.2 & 32.5 & \textbf{9.1} \\
E-n76-k7 & 41.7 & 46.6 & 33.7 & \textbf{7.2} & & \textbf{Average.} & 32.3 & 29.7 & 24.3 & \textbf{7.9} \\

\midrule
\multicolumn{11}{c}{\textbf{Set F}} \\
\cmidrule(lr){1-11}
F-n45-k4 & 41.0 & 62.7 & 45.9 & \textbf{7.0} & & F-n135-k7 & 39.9 & 61.4 & 28.9 & \textbf{22.0} \\
F-n72-k4 & 78.9 & 69.7 & 45.6 & \textbf{18.6} & & \textbf{Average.} & 53.3 & 64.6 & 40.1 & \textbf{15.9} \\

\midrule
\multicolumn{11}{c}{\textbf{Set M}} \\
\cmidrule(lr){1-11}
M-n101-k10 & 44.4 & 48.6 & 22.3 & \textbf{10.4} & & M-n200-k16 & 45.3 & 40.2 & 31.6 & \textbf{18.9} \\
M-n121-k7  & 38.0 & 47.6 & 22.9 & \textbf{22.3} & & M-n200-k17 & 46.2 & 35.7 & 35.4 & \textbf{13.7} \\
M-n151-k12 & 46.8 & 42.8 & 33.2 & \textbf{10.1} & & \textbf{Average.} & 44.2 & 43.0 & 29.1 & \textbf{15.1} \\

\midrule
\multicolumn{11}{c}{\textbf{Set P}} \\
\cmidrule(lr){1-11}
P-n16-k8 & 3.3 & 2.9 & 3.1 & \textbf{2.9} & & P-n51-k10 & 36.1 & 31.6 & 20.5 & \textbf{9.1} \\
P-n19-k2 & 14.3 & 30.7 & \textbf{10.2} & 21.7 & & P-n55-k7  & 29.8 & 30.4 & 17.2 & \textbf{4.9} \\
P-n20-k2 & 13.5 & 23.7 & \textbf{6.1} & 13.0 & & P-n55-k10 & 28.6 & 16.8 & 23.7 & \textbf{3.9} \\
P-n21-k2 & 20.8 & 27.2 & 12.5 & \textbf{3.6} & & P-n55-k15 & 28.3 & 13.8 & 10.3 & \textbf{2.7} \\
P-n22-k2 & 18.4 & 24.1 & 9.4 & \textbf{5.9} & & P-n60-k10 & 32.4 & 32.2 & 17.0 & \textbf{6.0} \\
P-n22-k8 & 28.7 & 34.7 & 20.8 & \textbf{4.4} & & P-n60-k15 & 38.2 & 30.8 & 14.3 & \textbf{11.2} \\
P-n23-k8 & 20.5 & 15.6 & 8.9 & \textbf{6.0} & & P-n65-k10 & 28.6 & 33.1 & 26.8 & \textbf{7.7} \\
P-n40-k5 & 19.7 & 23.4 & 21.5 & \textbf{4.0} & & P-n70-k10 & 29.2 & 32.6 & 14.3 & \textbf{8.0} \\
P-n45-k5 & 34.0 & 45.1 & 25.1 & \textbf{7.1} & & P-n76-k4  & 43.6 & 20.7 & 20.1 & \textbf{15.9} \\
P-n50-k7 & 30.0 & 19.8 & 21.8 & \textbf{8.2} & & P-n76-k5  & 26.0 & 27.8 & 21.8 & \textbf{9.9} \\
P-n50-k8 & 26.7 & 33.3 & 18.4 & \textbf{6.9} & & P-n101-k4 & 36.8 & 29.1 & 19.9 & \textbf{12.2} \\
P-n50-k10& 28.0 & 19.5 & 20.1 & \textbf{10.7} & & \textbf{Average.} & 26.8 & 26.0 & 16.7 & \textbf{8.1} \\

\bottomrule
\end{tabular}
\end{table*}


\begin{table*}[h]
\caption{Detailed optimality gaps (\%) on CVRPLib instances (Sets X). 
Results are grouped by dataset. The best results are highlighted in \textbf{bold}.}
\label{tab:cvrplib_efmps}
\centering
\small
\setlength{\tabcolsep}{3.5pt} 
\begin{tabular}{lrrrr clrrrr}
\toprule
\textbf{Instance} & \textbf{EoH} & \textbf{ReEvo} & \textbf{EoH-S} & \textbf{Ours} & \phantom{abc} & 
\textbf{Instance} & \textbf{EoH} & \textbf{ReEvo} & \textbf{EoH-S} & \textbf{Ours} \\
\midrule

\multicolumn{11}{c}{\textbf{Set X}} \\
\cmidrule(lr){1-11}
X-n101-k25 & 48.7 & 40.0 & 24.9 & \textbf{7.4} & & X-n204-k19 & 23.4 & 19.8 & 19.2 & \textbf{14.2} \\
X-n106-k14 & 7.7 & 10.8 & 6.5 & \textbf{3.6} & & X-n209-k16 & 20.4 & 17.0 & 11.7 & \textbf{8.1} \\
X-n110-k13 & 20.2 & 24.8 & 16.3 & \textbf{12.8} & & X-n214-k11 & 30.3 & 32.5 & 23.0 & \textbf{18.9} \\
X-n115-k10 & 50.8 & 46.3 & 39.4 & \textbf{11.3} & & X-n219-k73 & 1.8 & 48.9 & \textbf{1.4} & 7.1 \\
X-n120-k6  & 20.4 & 19.3 & 18.3 & \textbf{16.8} & & X-n223-k34 & 26.2 & 17.1 & 14.1 & \textbf{9.7} \\
X-n125-k30 & 16.0 & 21.0 & 14.2 & \textbf{7.4} & & X-n228-k23 & 38.5 & 26.9 & 26.2 & \textbf{11.9} \\
X-n129-k18 & 20.2 & 14.0 & 15.4 & \textbf{13.0} & & X-n233-k16 & 53.6 & 64.3 & 34.2 & \textbf{16.4} \\
X-n134-k13 & 61.0 & 50.8 & 24.6 & \textbf{13.0} & & X-n237-k14 & 23.1 & 19.0 & 18.3 & \textbf{16.6} \\
X-n139-k10 & 23.5 & 24.8 & 19.7 & \textbf{14.3} & & X-n242-k48 & 15.1 & 10.4 & 8.8 & \textbf{6.1} \\
X-n143-k7  & 53.5 & 43.8 & 35.4 & \textbf{13.5} & & X-n247-k50 & 35.8 & 33.2 & 29.7 & \textbf{9.2} \\
X-n148-k46 & 30.7 & 17.0 & 18.6 & \textbf{10.8} & & X-n251-k28 & 11.5 & 14.3 & 10.7 & \textbf{8.2} \\
X-n153-k22 & 44.5 & 39.0 & 28.7 & \textbf{10.0} & & X-n256-k16 & 19.9 & 17.2 & 15.0 & \textbf{8.4} \\
X-n157-k13 & 6.6 & 21.6 & 6.3 & \textbf{3.7} & & X-n261-k13 & 33.7 & 35.0 & 26.3 & \textbf{15.6} \\
X-n162-k11 & 12.7 & 24.4 & 17.5 & \textbf{9.5} & & X-n266-k58 & 9.2 & 10.7 & 8.0 & \textbf{7.4} \\
X-n167-k10 & 27.5 & 23.8 & 19.7 & \textbf{16.4} & & X-n270-k35 & 19.4 & 23.4 & 13.9 & \textbf{12.1} \\
X-n172-k51 & 43.4 & 40.8 & 35.9 & \textbf{14.2} & & X-n275-k28 & 13.8 & 21.4 & 11.7 & \textbf{5.9} \\
X-n176-k26 & 35.2 & 37.0 & 27.0 & \textbf{11.9} & & X-n280-k17 & 23.3 & 22.1 & 25.0 & \textbf{14.8} \\
X-n181-k23 & 7.5 & 15.6 & 6.6 & \textbf{2.5} & & X-n284-k15 & 27.7 & 23.6 & 20.7 & \textbf{16.7} \\
X-n186-k15 & 22.9 & 20.7 & 17.0 & \textbf{14.7} & & X-n289-k60 & 22.1 & 25.0 & 11.8 & \textbf{9.6} \\
X-n190-k8  & 17.0 & 14.9 & 18.2 & \textbf{8.2} & & X-n294-k50 & 50.7 & 36.4 & 22.1 & \textbf{17.4} \\
X-n195-k51 & 27.5 & 37.5 & 29.6 & \textbf{13.5} & & X-n298-k31 & 38.7 & 18.8 & 19.9 & \textbf{10.0} \\
X-n200-k36 & 18.3 & 16.9 & 10.4 & \textbf{7.7} & & \textbf{Average.} & 26.8 & 26.5 & 19.1 & \textbf{11.2} \\

\bottomrule
\end{tabular}
\end{table*}

\end{document}